\documentclass{article} %
\usepackage{iclr2027_conference,times}

\iclrfinalcopy

\usepackage[utf8]{inputenc} %
\usepackage[T1]{fontenc}    %
\usepackage{hyperref}       %
\usepackage{url}            %
\usepackage{booktabs}       %
\usepackage{amsmath}        %
\usepackage{amsfonts}       %
\usepackage{nicefrac}       %
\usepackage{microtype}      %
\usepackage{xcolor}         %
\usepackage{graphicx}       %
\usepackage{subcaption}     %
\usepackage{tikz}           %
\usetikzlibrary{positioning}
\usepackage{algorithm}      %
\usepackage{algpseudocode}  %
\usepackage{multirow}
\usepackage{wrapfig}

\definecolor{linkcolor}{RGB}{84,84,192}
\hypersetup{
  colorlinks=true,
  linkcolor=linkcolor,
  citecolor=linkcolor
}

\makeatletter
\@namedef{HyRef@apx@type@section}{}
\@namedef{HyRef@apx@type@subsection}{}
\@namedef{HyRef@apx@type@subsubsection}{}
\@namedef{HyRef@apx@type@section*}{}
\@namedef{HyRef@apx@type@subsection*}{}
\@namedef{HyRef@apx@type@subsubsection*}{}
\def\HyRef@autoref#1#2{%
  \begingroup
    \Hy@safe@activestrue
    \HyRef@apx@getnum{#2}%
    \expandafter\HyRef@autosetref\csname r@#2\endcsname{#2}{#1}%
  \endgroup
}
\def\HyRef@apx@getnum#1{%
  \ltx@IfUndefined{r@#1}%
    {\let\HyRef@apx@num\@empty}%
    {\expandafter\expandafter\expandafter\HyRef@apx@@getnum
       \csname r@#1\endcsname\@empty\@empty\@empty\@nil}%
}
\def\HyRef@apx@@getnum#1#2#3#4#5\@nil{\def\HyRef@apx@num{#1}}
\let\HyRef@apx@num\@empty
\let\HyRef@orig@testreftype\HyRef@testreftype
\def\HyRef@testreftype#1.#2\\{%
  \def\HyRef@apx@args{#1.#2\\}%
  \@ifundefined{HyRef@apx@type@#1}%
    {\HyRef@orig@testreftype#1.#2\\}%
    {\expandafter\HyRef@apx@testnum\HyRef@apx@num.\@nil}%
}
\def\HyRef@apx@testnum#1#2\@nil{%
  \ifcat A\noexpand#1%
    \edef\HyRef@currentHtag{\noexpand\appendixautorefname\noexpand~}%
  \else
    \expandafter\HyRef@orig@testreftype\HyRef@apx@args
  \fi
}
\makeatother
\definecolor{maegreen}{HTML}{3C7E64}
\definecolor{jepaorange}{HTML}{A44F39}
\newcommand{\maegreen}[1]{\textcolor{maegreen}{#1}}
\newcommand{\jepaorange}[1]{\textcolor{jepaorange}{#1}}

\title{What masking geometry works best for EEG foundation models?\\
{\Large A controlled evaluation across MAE and JEPA}}

\author{%
  Pierre Guetschel \\
  Donders Institute for Brain, Cognition and Behaviour\\
  Radboud University\\
  Nijmegen, The Netherlands \\
  \texttt{pierre.guetschel@donders.ru.nl} \\
  \And
  Bruno Aristimunha\\
  Yneuro,\\
  University of California San Diego,\\
  Paris, France \\
  \texttt{b.aristimunha@gmail.com}\\
  \And
  Yassine El Ouahidi \\
  Lab-STICC, IMT Atlantique\\
  Brest, France \\
  \texttt{yassine.elouahidi@mistral.ai} \\
  \And
  Arnaud Delorme \\
  SCCN, INC, SDSC\\
  University of California San Diego, USA\\
  CNRS, France\\
  \texttt{adelorme@ucsd.edu} \\
  \And
  Thomas Moreau\thanks{These authors jointly supervised this work.} \\
  Universit\'e Paris-Saclay, Inria, CEA\\
  Palaiseau, France \\
  \texttt{thomas.moreau@inria.fr} \\
  \And
  Michael Tangermann\footnotemark[\value{footnote}] \\
  Donders Institute for Brain, Cognition and Behaviour\\
  Radboud University\\
  Nijmegen, The Netherlands \\
  \texttt{michael.tangermann@donders.ru.nl} \\
}

\begin{document}

\maketitle
\lhead{Preprint.} %

\begin{abstract}

  \looseness=-1
  EEG foundation models hold promise for scalable brain-signal decoding across clinical and cognitive neuroscience applications, yet their pre-training pipelines remain poorly understood. Among design choices, the masking strategy is particularly critical: it determines what the network must predict and from which context.
  Yet it has never been ablated in isolation, as each new model bundles a new masking strategy with a new backbone and objective.
  In this paper, we formalize the design choices for spatio-temporal masking strategies and train
  various models with a single pipeline under varying masking configurations across two SSL frameworks (MAE and JEPA).
  We then systematically evaluate the resulting 58 pre-trained models on the 12 datasets of OpenEEGBench under a linear probe.
  Both frameworks agree on an optimal masking configuration and on shared failure modes.
  Outside these, performance is robust: 11 MAE and 9 JEPA configurations are statistically indistinguishable from the best. We further identify a novel JEPA-specific failure mode, tagged \emph{bias-inflation collapse}, invisible to standard detectors.
  With a well-chosen mask, our pipeline reaches REVE-level downstream performance at a fraction of REVE's pre-training compute.

\end{abstract}

\section{Introduction}

Self-supervised foundation models for electroencephalography (EEG) have multiplied rapidly in the past few years.
Models such as REVE~\citep{ouahidiREVEFoundationModel2025}, LaBraM~\citep{jiangLargeBrainModel2024a}, BIOT~\citep{yangBIOTBiosignalTransformer2023}, BrainBERT~\citep{wangBrainBERTSelfsupervisedRepresentation2023a}, and Neuro-GPT~\citep{cuiNeuroGPTFoundationModel2024} are reporting state-of-the-art performance on batteries of downstream decoding tasks, ranging from clinical applications such as abnormality detection~\citep{obeid2016tuh} or sleep staging~\citep{khalighiISRUCSleepComprehensivePublic2016} to cognitive neuroscience such as motor imagery~\citep{brunner2008bcic2a} or emotion recognition~\citep{liu2022seedv}.
The large majority of these models share a common design principle: a masked-prediction pretext task, in which a portion of the input is hidden and the network is trained to recover it from the visible context.
To define this task, a critical ingredient is the choice of the masking strategy, which defines what is the context and what is the target in this predictive task.
The literature uses a remarkably diverse set of strategies: random patch masking, whole-channel masking, contiguous time-window masking, and various spatio-temporal block schemes.
Yet every new model bundles together multiple design changes at once, rarely ablating each choice in isolation: a different backbone, a different pre-training corpus, a different pretext objective, and---crucially---a different masking strategy.
This makes it impossible to attribute downstream gains to any single component.

In adjacent self-supervised fields for other modalities, the masking strategy has long been recognised as a crucial design choice: MAE~\citep{heMaskedAutoencodersAre2022} ablates the masking ratio, I-JEPA~\citep{assranSelfSupervisedLearningImages2023a} and V-JEPA~\citep{bardesRevisitingFeaturePrediction2024} ablate multi-block masking shapes and counts, and ST-MAE~\citep{feichtenhoferMaskedAutoencodersSpatiotemporal2022} establishes a spatio-temporal masking precedent for video.
As no comparable study exists for EEG, the community lacks principled guidance on how to apply masking.

This motivates the central question of this paper: \emph{What masking geometry works best for EEG foundation models?} and \emph{What's its impact on their downstream performance?}
A robust answer may have a double impact:
First, it provides an actionable default for practitioners regarding an important design choice that is currently made by inheritance or guesswork.
Second, it sheds light on the deeper question of what kind of supervision a mask provides for EEG data, and how that supervision varies with the mask's spatial and temporal structure.

We approach this question with a controlled grid sweep.
We first introduce a formal framework that unifies the masking strategies of the EEG self-supervised literature under a single formalism with three parameters: a spatial radius~$r$, a temporal length~$L$, and a target masking ratio~$\rho$.
Tuning $r$ and $L$ continuously interpolates between random patch masking, whole-channel masking, contiguous time-window masking, and spatio-temporal block masking.
We sweep a parameter grid of $29$ $(L, r)$ configurations covering $6$ lengths and $5$ radii (excluding the degenerate full-mask cell). Following the results of REVE~\citep{ouahidiREVEFoundationModel2025}, we keep the mask ratio fixed at $\rho=0.55$.
To assess whether the conclusions generalise beyond a single pretext, we replicate the entire sweep for two SSL frameworks: input-space reconstruction à la MAE~\citep{heMaskedAutoencodersAre2022} and latent-space prediction à la JEPA~\citep{assranSelfSupervisedLearningImages2023a}.
This yields $58$ pre-trained foundation models, all sharing the REVE pipeline except for the two axes under study. Each model is evaluated on OpenEEGBench (OEB)~\citep{guetschelOpenEEGBench2026} using a linear probe on frozen features.

\looseness=-1
Our findings illuminate four points.
First, MAE and JEPA agree on which mask geometries help and which do not: a shared optimum at moderate spatial radius and short temporal length, and three shared failure modes (single-channel blocks, full-channel blocks, and most long temporal blocks).
Second, within the working regime, downstream performance is largely insensitive to the precise mask choice --- many configurations are statistically indistinguishable from the best.
Third, JEPA admits a multichannel-specific failure mode that we name \emph{bias-inflation collapse} and which standard collapse detectors miss.
Fourth, our pipeline is compute-efficient: at the optimum, both frameworks reach REVE-level downstream performance for a fraction of the original pre-training cost.

We contribute (i)~a formal three-parameter framework that unifies the masking strategies of the EEG self-supervised literature; (ii)~$58$ open-source pre-trained checkpoints spanning the full $(L, r)$ grid under both MAE and JEPA pretexts; (iii)~an actionable default mask for practitioners pre-training EEG foundation models; and (iv)~the characterisation of \emph{bias-inflation collapse}, a JEPA-specific failure mode on multichannel signals that we do not find documented in the prior literature.
In the tradition of design-axis ablations such as \citet{tianWhatMakesGood2020}'s ``What makes for good views for contrastive learning?'', we isolate one component of the pipeline rather than introducing a new model.
The code to reproduce the pre-training and evaluation pipelines, as well as the pre-trained checkpoints, is available at 
\url{https://pierregtch.github.io/eeg-fm-masking}. 

\section{Related work}

\paragraph{EEG foundation models.}
Self-supervised foundation models for EEG and broader biosignals~\citep{leeComprehensiveReviewBiosignal,zhouBrainFoundationModels2025} increasingly rely on masked-prediction pretexts, but their gains are usually reported at the level of a full pipeline.
Recent models change tokenisation~\citep{jiangLargeBrainModel2024a,yangBIOTBiosignalTransformer2023}, attention or channel aggregation~\citep{wangCBraModCrissCrossBrain2025,zhouCSBrainCrossscaleSpatiotemporal2025,donerLUNAEfficientTopologyAgnostic2025}, the pretext objective~\citep{kostasBENDRUsingTransformers2021a,cuiNeuroGPTFoundationModel2024,wangEEGPTPretrainedTransformer2024}, and even the recording regime~\citep{wangBrainBERTSelfsupervisedRepresentation2023a}.
This diversity makes it difficult to tell which design choices are responsible for downstream improvements, and motivates controlled studies that isolate one axis at a time.

Closest to our setting, REVE~\citep{ouahidiREVEFoundationModel2025} pairs a vanilla transformer with an MAE objective and a mixed spatio-temporal block mask at $\rho{=}0.55$; ST-EEGFormer~\citep{yangLearningRobustEEG2024} similarly shows that a plain MAE on raw patches can compete with more elaborate EEG architectures.
Whereas these works propose complete pipelines with a fixed mask, we keep the REVE pipeline as a reference point and ablate only the mask geometry: a controlled sweep of $29$ $(L, r)$ configurations, replicated under both MAE and JEPA, yielding $58$ checkpoints that map shared optima, shared failure modes, and a JEPA-specific collapse.

\paragraph{Masking strategies for sequential and multichannel signals.}
Block-masked prediction has a clear lineage from BERT~\citep{devlinNoTitleFound2019} for language, through MAE~\citep{heMaskedAutoencodersAre2022} for images, to ST-MAE~\citep{feichtenhoferMaskedAutoencodersSpatiotemporal2022} for video, where the choice of mask shape is recognised as a primary design axis.
EEG masked models also vary their masks, but usually as a fixed design choice rather than as an object of study: Signal-JEPA~\citep{guetschelSJEPASeamlessCrossdataset2024} uses channel blocks, EEG-VJEPA~\citep{hojjatiVideoEEGAdapting2026} adapts video-JEPA spatio-temporal tubelets, EEGPT~\citep{wangEEGPTPretrainedTransformer2024} uses a high-radius mixed mask, CBraMod~\citep{wangCBraModCrissCrossBrain2025} masks patches independently at $\rho{\approx}0.5$, EEG2Rep~\citep{mohammadifoumaniEEG2RepEnhancingSelfsupervised2024} preserves informative subsequences while predicting latent targets, and MAEEG~\citep{chienMAEEGMaskedAutoencoder2022a} reports that concentrated temporal masks improve downstream sleep staging.
These choices cover single-channel, full-channel, temporal, and spatio-temporal regimes, but the EEG literature lacks a common parameterization that shows how these regimes relate to one another or how performance changes between them.
Our $(L, r)$ parameterization recovers these shapes as boundary cases and indexes a continuous interpolation between them under a fixed pipeline.

\paragraph{MAE vs.~JEPA beyond EEG.}
The contrast between input-space reconstruction and latent-space prediction has been studied at length outside EEG, in vision~\citep{heMaskedAutoencodersAre2022,assranSelfSupervisedLearningImages2023a}, video~\citep{bardesRevisitingFeaturePrediction2024,feichtenhoferMaskedAutoencodersSpatiotemporal2022}, and general-purpose SSL~\citep{baevskiData2vecGeneralFramework2022a,baevskiEfficientSelfsupervisedLearning2022a}.
Closest to our work, \citet{vanasselJointEmbeddingVs2025} give a closed-form characterization of when latent-space prediction outperforms pixel-space reconstruction under high-variance observation noise, a regime that arguably matches EEG.
While their analysis is theoretical and based on synthetic Gaussian noise, we measure the MAE--JEPA contrast empirically across $12$ real EEG datasets and $58$ pre-trained checkpoints under a fixed pipeline.

\paragraph{Ablation-centric studies.}
\looseness=-1
Several ablation-centric studies have shaped SSL and transformer practice without introducing a new architecture: \citet{tianWhatMakesGood2020} characterized what makes a good view for contrastive learning, \citet{steinerHowTrainYour2021} disentangled pipeline choices for vision transformers, and \citet{zhaiScalingVisionTransformers2022} mapped scaling behavior for vision transformers.
Our work follows this tradition in EEG: rather than proposing another foundation model, it isolates masking geometry as a critical pre-training axis and provides the controlled ablation that is missing from the current EEG literature.

\section{Controlled evaluation of masking geometry and SSL framework}\label{sec:method}

We propose a controlled experimental design in which $58$ EEG foundation models are pre-trained and evaluated under a single fixed recipe, varying only two axes: the SSL framework (MAE vs.\ JEPA) and the spatio-temporal masking geometry $(L,r)$. Any difference in downstream performance can therefore be attributed to one of these two axes or their interaction.

\subsection{A shared pre-training pipeline}
\label{sub:training}

\begin{figure}[t]
  \centering
  \begin{tikzpicture}[every node/.style={inner sep=0pt, outer sep=0pt}]
    \node (a) {\includegraphics[width=\textwidth]{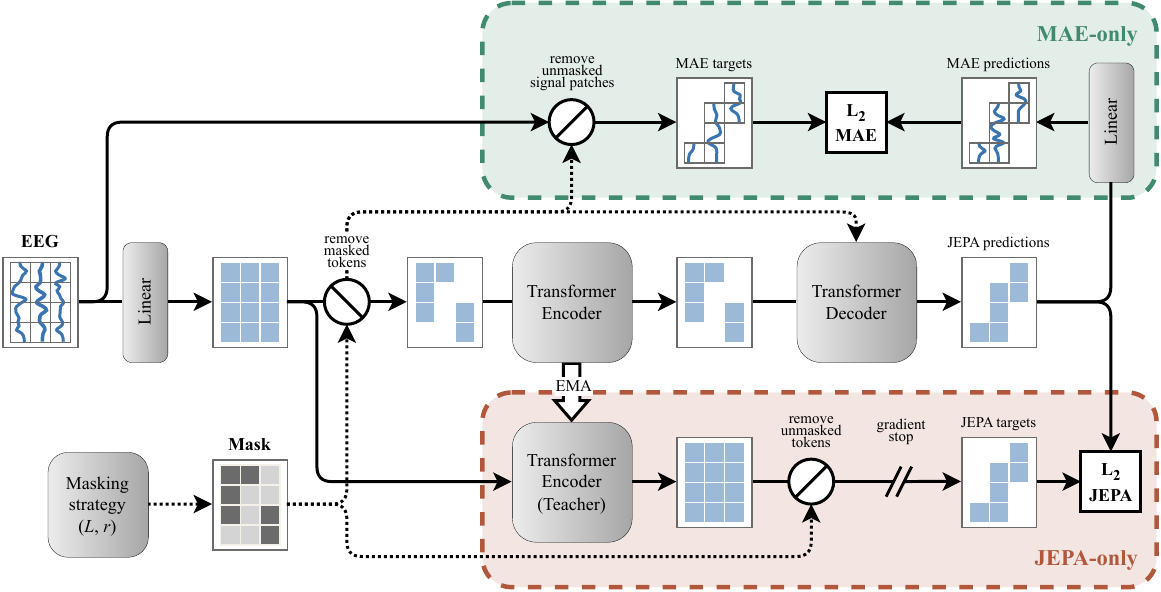}};
    \node[below=0.5cm of a] (b) {\includegraphics[width=\textwidth]{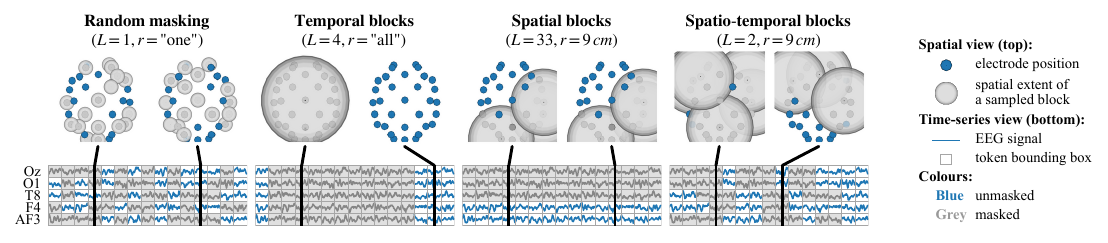}};
    \node[anchor=north west, font=\Large\bfseries] at (a.north west) {A};
    \node[anchor=north west, font=\Large\bfseries] at (b.north west) {B};
  \end{tikzpicture}
  \caption{
    Controlled cross of two design axes for EEG masked-prediction pre-training.
    \textbf{(A)~Framework axis.} The \maegreen{MAE} and \jepaorange{JEPA} pretexts are superimposed on a shared pipeline; components outside the coloured dashed boxes (linear tokeniser, masking, encoder, decoder) are shared between the two, while each pretext adds its own branch (\maegreen{green: MAE input-space reconstruction}; \jepaorange{orange: JEPA latent prediction against an EMA teacher}).
    \textbf{(B)~Masking axis.} Block-masks parameterised by spatial radius~$r$, temporal length~$L$, and target ratio~$\rho$.
    This formalism unifies the masking strategies used in the EEG self-supervised literature, recovering four canonical regimes at extreme $(L, r)$ values: random patches, pure temporal blocks, pure spatial blocks, and spatio-temporal blocks.
    We sweep $29$ $(L, r)$ configurations at fixed $\rho=0.55$.
    Holding every other component fixed (backbone, optimiser, corpus, tokenisation, downstream protocol) yields the $58 = 2 \times 29$ pre-trained foundation models that we evaluate on OpenEEGBench under a linear probe on frozen features.
  }
  \label{fig:method}
\end{figure}

\paragraph{A shared training pipeline.}
First, we fix the backbone shared across all our foundation models (FMs) to the \emph{REVE-Small} variant of REVE~\citep{ouahidiREVEFoundationModel2025}.
We choose REVE because three benchmarks, OpenEEGBench~\citep{guetschelOpenEEGBench2026}, NeuralBench~\citep{banvilleNeuralBenchUnifyingFramework2026} and NeuroAtlas~\citep{kontras2026neuroatlasbenchmarkingfoundationmodels}, independently identified it as the strongest published EEG foundation model.
We nevertheless test the robustness of our results to the backbone in \autoref{app:ablation-backbones}.
The REVE architecture consists of a transformer with $4$ encoder layers, $2$ decoder layers, $8$ attention heads, and a token dimension of $512$, for a total of $\sim\!12.7\,$M parameters, excluding the decoder which is unused for downstream evaluations.
This architecture uses GEGLU feed-forward blocks~\citep{shazeerGLUVariantsImprove2020}, RMSNorm~\citep{zhangRootMeanSquare2019}, the LLaMA-style FFN arrangement~\citep{touvronLLaMAOpenEfficient2023}, and a fixed 4D sinusoidal positional encoding (see \autoref{app:reve-small-arch} for a complete description).
Then, we fix the pre-training dataset for all our FM trainings to use the open part of the corpus used to train REVE.
REVE is trained on a corpus of approximately $6\,$TB, which bundles a mix of open-licence datasets and access-restricted datasets. We restrict ourselves to the open-licence subset consisting of $4.4\,$TB so that we can freely distribute the resulting pre-trained checkpoints and training scripts.
We also fix the optimization procedure for all our training, using AdamW~\citep{loshchilovDecoupledWeightDecay2017} with a one-epoch warmup, a one-cycle cosine learning-rate schedule, and bf16 mixed precision (see \autoref{app:training-hyperparams} for all training hyperparameters).
The input of the network is fixed to $30\,$s windows with $32$ channels.
The training input windows are non-overlapping windows from the whole open corpus, rescaled with a robust scaler whose statistics are estimated only on the current window, and clipped to avoid values exceeding 15 standard deviations.
We randomly sample $32$ channels, padding with zeros when fewer channels are present.
Windows are tokenised identically across configurations, with tokens of $1\,$s length and $0.1\,$s overlap.
Each configuration is pre-trained on the same fixed number of seen tokens, corresponding to $10$ epochs over the corpus.
Additional details on this shared pipeline are given in \autoref{app:frameworks-shared-details}.
We verified in preliminary experiments that our pipeline reproduces results close to those of REVE before launching the full ablation study.

\paragraph{The Framework axis.}
For reconstruction-based self-supervised learning, two main frameworks are competing.
On one side, Masked Autoencoders (\emph{MAE}) reconstruct masked input patches in the input space from the visible (unmasked) tokens, under an $L_2$ reconstruction loss.
Typically, a final linear layer projects the decoded output tokens back to the input space.
This framework is relatively straightforward and therefore uses neither a secondary loss nor a teacher network.
On the other side, Joint-Embedding Predictive Architectures (\emph{JEPA}) replace input-space reconstruction with a latent-space prediction: both the context and the target are first embedded into a latent space, and the target embeddings are then predicted from the context embeddings.
To provide stable targets during training, the target embeddings are produced by an Exponential Moving Average (EMA) teacher that receives the full (unmasked) input and outputs the embeddings of the masked tokens. These target embeddings are predicted from the embeddings of the unmasked tokens under an $L_2$ loss.
To stay in the simplest setting and vary as little as possible across configurations, we deliberately add no secondary loss to stabilise training.
Although JEPA variants often add explicit anti-collapse terms such as VICReg-style variance--covariance penalties~\citep{bardesVICRegVarianceInvarianceCovarianceRegularization2021}, C-JEPA's contrastive correction~\citep{moConnectingJointEmbeddingPredictive2024}, or SIGReg in LeJEPA~\citep{balestrieroLeJEPAProvableScalable2025}, we treat these as separate pipeline choices rather than part of the framework axis.
This keeps JEPA at strict parity with MAE on the ``no auxiliary loss'' axis.
Preliminary runs showed no representational collapse at the REVE scale under this setting, so we avoided adding anti-collapse regularisation to keep the comparison as controlled as possible.
These two frameworks are illustrated in \autoref{fig:method}~(A), with additional details given in Appendices~\ref{app:mae-details} and~\ref{app:jepa-details}.

\paragraph{The Masking axis.}
Here, we unify the various masking strategies proposed in the literature into a single parametric family.
As EEG signals are spatio-temporal signals, with auto-correlation in both space and time, the definition of masking strategies needs to account for these two directions.
Let $x \in \mathbb{R}^{N_c \times N_t}$ denote a tokenized EEG segment with $N_c$ (non-padding) channels at scalp positions $\{p_c\}_{c=1}^{N_c} \subset \mathbb{R}^3$ and $N_t$ temporal patches.
A binary mask $M \in \{0,1\}^{N_c \times N_t}$ is defined based on a set of $K$ blocks $(B_1, \dots, B_K)$ parameterized by a spatial center and temporal start $(c_i, t_i)$, associated with a spatial radius~$r$ and a temporal length~$L$.
This parameterization controls the span of the mask in both directions through the dimensions of the block:
\begin{align}
  B_i = \bigl\{\, (c, t) \;:\; \|p_c - p_{c_i}\|_2 \le r \ \text{ and }\ t_i \le t \le t_i + L - 1 \,\bigr\}.
\end{align}
The final mask is defined by $M_{c,t} = 1$ when $(c, t) \in \bigcup_{i=1}^K B_i$.
The mask is then applied only to non-padding channels (at most $32$).

To select the blocks $B_i$, each block is sampled independently from the others by sampling $(c_i, t_i)$ uniformly, with $c_i$ drawn uniformly over channels and $t_i \sim \mathrm{Unif}\{1, \dots, N_t - L + 1\}$.
The number of blocks $K$ is chosen such that the expected mask ratio equals a fixed target $\rho^\star$, while accounting for overlaps between blocks.
Since the $K$ blocks are sampled i.i.d., a fixed cell $(c, t)$ is left unmasked with probability $(1-f)^K$, where $f \in [0,1]$ denotes the expected fraction of cells covered by a single block.
Setting $\mathbb{E}[\rho] = 1 - (1-f)^K$ equal to $\rho^\star$ and rounding up gives:
\begin{align}\label{eq:K}
  K \;=\; \biggl\lceil \frac{\log(1 - \rho^\star)}{\log(1 - f)} \biggr\rceil,
  \qquad
  f \;=\; f_\mathrm{sp}(r) \cdot \frac{L}{N_t},
\end{align}
where $f_\mathrm{sp}(r)$ is the expected fraction of channels within a distance $r$ of a uniformly drawn center channel.
We use a uniform-scalp approximation $f_\mathrm{sp}(r) = \pi r^2 / A_\mathrm{scalp}$, with $A_\mathrm{scalp}$ the scalp surface area.
The full sampling procedure is given in Algorithm~\ref{alg:mask-maker} in \autoref{app:mask-algorithm}.
In the following, we fix $\rho^\star = 0.55$ based on the pre-training performance obtained on initial training of the REVE architecture~\citep{ouahidiREVEFoundationModel2025}; \autoref{app:ablation-rho} explains why $\rho$ cannot be varied jointly with $(L, r)$ and shows that our conclusions hold across ratios.

The two extreme values of $r$ are defined as special cases with exact closed forms: $f_\mathrm{sp} = 1/N_c$ when $r$ is below the smallest inter-channel distance (each block covers a single channel), and $f_\mathrm{sp} = 1$ when $r$ exceeds the largest inter-channel distance (each block covers every channel).
We denote these two special cases as $r=\texttt{"one"}$ and $r=\texttt{"all"}$, respectively.

This masking strategy yields spatio-temporal masks, similar to the block-masking strategy used in REVE~\citep{ouahidiREVEFoundationModel2025} and Signal-JEPA~\citep{guetschelSJEPASeamlessCrossdataset2024}.
Particular values of $r$ and $L$ recover other strategies: using $L=1, r=\texttt{"one"}$ yields the random masking strategy~\citep{wangCBraModCrissCrossBrain2025}, while $r=\texttt{"all"}$ with short blocks yields the temporal block masking, a strategy that forces the model to learn to predict a full time-segment. Finally, $L = 33$ yields spatial blocks, where the goal of the model is to reconstruct a complete channel from the information of other channels.

\subsection{Downstream evaluation protocol}\label{sub:downstream}

\paragraph{Downstream protocol.}
To evaluate the obtained foundation models, we consider their performance on a set of diverse downstream tasks using only linear probing.
A core promise of FMs is to produce rich embeddings of EEG recordings, which is best measured with a simple linear model on top of them, as it shows how well the model disentangles complex information.
This protocol also maximizes the number of downstream datasets that can be evaluated under a fixed compute budget while still providing a strong signal of representation quality.
To this aim, our evaluation relies on OpenEEGBench~\citep{guetschelOpenEEGBench2026}, an open-source, community-driven benchmark to evaluate EEG-FM, originally introduced in \citet{guetschelOpenEEGBenchLiveCommunityDriven}.\ificlrfinal\footnote{OpenEEGBench is co-developed by some of the authors of this paper.}\fi
All pre-trained checkpoints are evaluated with a ridge-regression linear probe fitted on frozen representations, on the $12$ OpenEEGBench datasets.
We strictly follow the train/validation/test splits provided by the benchmark.
Final models (epoch 10) are evaluated across 5 seeds that vary the random projection (the only stochastic component of the foundation-model downstream pipeline).
The per-dataset feature dimensions before the probe's random projection are reported in \autoref{app:oeb-feature-dims}, and per-dataset class distributions in \autoref{app:oeb-class-distributions}. Additional details on the downstream protocol are given in \autoref{app:downstream-details}.

\paragraph{Score aggregation and statistical testing.}
To aggregate the results for each model, raw scores are first normalised within the dataset (see \autoref{app:score-normalisation} for the exact per-dataset normalisation) and then aggregated as the mean of the normalised scores over the datasets and the seeds.
Raw per-dataset scores are reported in \autoref{app:per-dataset-breakdown}.
Using the repeated evaluation across seeds and datasets, we also perform statistical testing using a two-level bootstrap that propagates the variance at both levels.
This assesses if a configuration is better than the others, in the spirit of \citet{demsarStatisticalComparisonsClassifiers2006} and \citet{agarwalDeepReinforcementLearning2021a}.
At each of the $T = 10\,000$ bootstrap replicates, the $12$ OpenEEGBench datasets are resampled with replacement (treating the benchmark as a sample of plausible EEG datasets), and within each resampled dataset, the $5$ seeds of every configuration are themselves resampled with replacement, independently across configurations.
Because raw scores are not comparable across datasets, we rank configurations within each dataset before averaging the $12$ within-dataset ranks into a single mean rank $\bar R_c^{(t)} \in [1, 29]$ per configuration $c$.
We summarize the resulting joint distribution by the pairwise dominance probability $\mathbb{P}(c \succ c') = \frac{1}{T}\sum_t \mathbf{1}[\bar R_c^{(t)} < \bar R_{c'}^{(t)}]$. We choose $T = 10\,000$ so that the Monte Carlo standard error of $\mathbb{P}(c \succ c')$ stays below $0.005$ for any underlying value, comfortably above the $T \geq 2\,000$ recommended by~\citet{agarwalDeepReinforcementLearning2021a}. We call configuration $c$ \emph{strictly better than} $c'$ when $\mathbb{P}(c \succ c') > 0.975$; pairs with $\mathbb{P} \in [0.025, 0.975]$ are reported as equally ranked.

\paragraph{Baselines.}
The performance of our pre-trained models is compared against two families of baselines, both evaluated within the OpenEEGBench framework.
First, we compare our models to the \emph{REVE baseline}, which corresponds to the REVE architecture with pre-trained weights from the REVE-Base checkpoint ($69\,$M parameters, pre-trained on the full $6\,$TB REVE corpus) released on HuggingFace~\citep{ouahidiREVEFoundationModel2025}.
This model is evaluated under the same ridge-regression linear probe on frozen features as our models.
A comparison with other published EEG foundation models is given in \autoref{app:sota-comparison}.
Additionally, we report the performances of three non-foundation baselines, chosen to span state-of-the-art models for the various tasks included in OpenEEGBench.
We include EEGNet~\citep{lawhernEEGNetCompactConvolutional2018}, EEGConformer~\citep{songEEGConformerConvolutional2023a}, and ShallowFBCSPNet~\citep{schirrmeisterDeepLearningConvolutional2017a}, using their Braindecode~\citep{aristimunhaBraindecodeToolboxDecoding2026} reference implementations.
These models are also trained and evaluated within OpenEEGBench under a full per-task training protocol.
Here, our evaluation departs from the evaluation of FMs due to the nature of these supervised baselines: our goal is not to investigate the quality of the learned representations, but to provide a reference point for each task.

\subsection{Grid sweep design}

We pre-train multiple EEG foundation models that span the Cartesian product of the two reconstruction-based SSL frameworks (MAE and JEPA) and a grid of spatio-temporal masking configurations parameterised by a temporal length~$L$ and a spatial radius~$r$.
The grid spans $r \in \{\texttt{"one"}, 6, 9, 12, \texttt{"all"}\}\,$cm and $L \in \{1, 2, 4, 8, 16, 33\}$ patches, yielding $30$ configurations. We remove the degenerate $(L=33, r=\texttt{"all"})$ where all the signal is masked, resulting in a total of $29$ configurations per framework.
The per-cell number of blocks~$K$ is listed in \autoref{tab:K-grid} of \autoref{app:mask-calibration}, where we also verify empirically that the actual mask ratio matches $\rho^\star = 0.55$ across the $29$ cells.

\section{Masking geometry drives downstream performance}\label{sec:results}

\begin{wrapfigure}[26]{r}{.5\textwidth}
  \centering
  \vskip-1.5em
    \includegraphics[width=\linewidth]{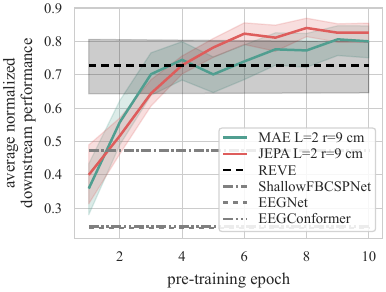}
    \caption{Mean normalised downstream performance on OpenEEGBench across the $12$ datasets and $5$ probe seeds, as a function of the pre-training epoch, for both frameworks at the optimal mask $(L, r) = (2\,\text{s}, 9\,\text{cm})$.
    Shaded bands are the $95\%$ confidence interval across seeds and datasets.
    The dashed black line locates the REVE baseline; horizontal dashed grey lines locate the three supervised baselines (EEGNet, EEGConformer, ShallowFBCSPNet) trained end-to-end per task.
  Both frameworks reach the REVE baseline level by epoch~$\sim\!4$ and remain at or above it.}
    \label{fig:oeb-best}
\end{wrapfigure}

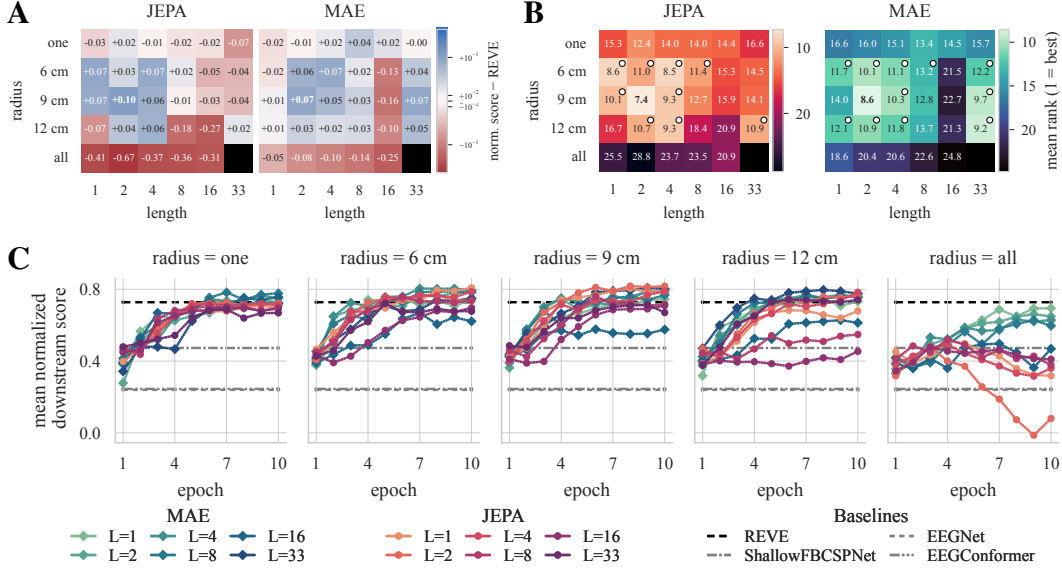
\begin{figure}[t]
  \centering
  \begin{tikzpicture}[every node/.style={inner sep=0pt, outer sep=0pt}]
    \node[anchor=north west] (a) at (0,0) {\includegraphics[height=2.95cm]{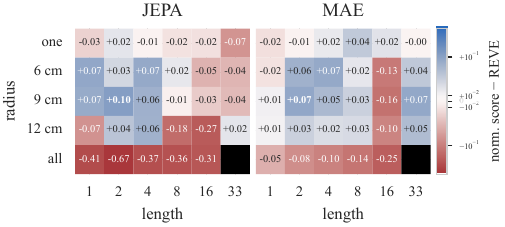}};
    \node[anchor=north east] (b) at (\textwidth,0) {\includegraphics[height=2.95cm]{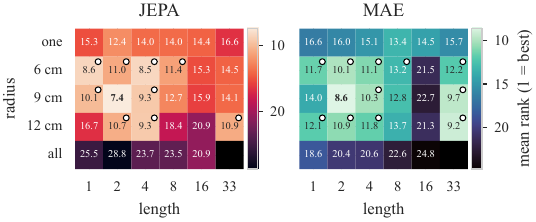}};
    \node[anchor=north east] (c) at ([yshift=-0.3cm]b.south east) {\includegraphics[width=.98\textwidth]{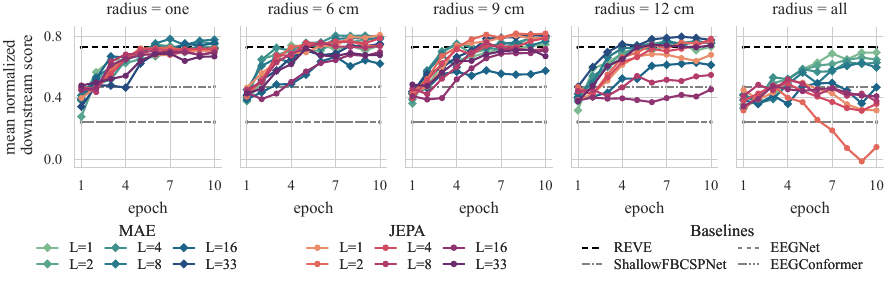}};
    \node[anchor=north west, font=\large\bfseries] at (a.north west) {A};
    \node[anchor=north west, font=\large\bfseries] at (b.north west) {B};
    \node[anchor=north west, font=\large\bfseries] at ([xshift=-0.02\textwidth]c.north west) {C};
  \end{tikzpicture}
  \caption{
    Effect of the masking geometry $(L, r)$ on downstream performances.
    \textbf{(A)~Mean normalised score} above the REVE baseline ($0 =$ the public REVE-Base checkpoint) across the $12$ OpenEEGBench datasets~$\times$~$5$ probe seeds.
     \textbf{(B)~Mean rank} ($1=$ best) per cell under the hierarchical bootstrap ($T=10{,}000$ replicates over datasets and seeds). The bold cell in each framework is the bootstrap winner; white circles mark cells not statistically distinguishable from it at $\mathbb{P}(\text{winner}\!\succ\! c)\le 0.975$ (ex-aequo cluster).
    \textbf{(C)~Training trajectories:} score of (A) as a function of the pre-training epoch, faceted by spatial radius~$r$, for both frameworks and all six values of $L$. Dashed lines locate the REVE baseline (black) and the three supervised baselines (grey). All foundation-model curves use a frozen-feature ridge probe; supervised baselines are trained end-to-end per task.
  }
  \label{fig:oeb-grid}
\end{figure}

We investigate the downstream performance of the 58 pre-trained models of the core grid and uncover the central role of spatio-temporal masking geometry. We first report headline results at the optimal mask configuration (\autoref{fig:oeb-best}), establishing that, with the right mask, our pipeline learns embeddings on par with the REVE baseline under both frameworks, at a fraction of its pre-training cost. We then open the box: \autoref{fig:oeb-grid} maps the full $(L,r)$ grid and reveals which configurations consistently win or fail, and investigate why, anchored by the spatial-vs.-temporal redundancy structure of EEG. Per-dataset breakdowns are in \autoref{app:per-dataset-breakdown}.

\paragraph{Headline at the optimal mask.}
\autoref{fig:oeb-best} previews the main outcome of this section.
At the optimal mask geometry $(L, r) = (2\,\text{s}, 9\,\text{cm})$, both embeddings learned with MAE and JEPA reach REVE-baseline level within a few pre-training epochs and stabilize at or above it throughout the rest of the training.
These performances are obtained with fewer parameters and a fraction of the compute used to pre-train the REVE baseline.
Choosing a suitable mask is therefore sufficient to match REVE-Base-level performance under both pretexts; the rest of this section unpacks how we arrive at this optimum, why this mask works, and where the approach breaks.
The supervised baselines (EEGNet, EEGConformer, ShallowFBCSPNet) are trained end-to-end on each downstream task and are shown only as a per-task reference for what task-specific supervised training achieves; they are not directly comparable to the foundation-model curves, which use a frozen linear probe.

\paragraph{Spatial vs.\ temporal redundancy.}\label{par:spat-temp-redundancy}
The two axes along which we mask reveal very different structures.
The within-channel autocorrelation in the EEG drops sharply with increased time-lag, falling below $R^2 \approx 0.2$ within $\sim\!2\,$s, whereas between-channel correlation decays much more slowly with electrode distance and remains non-negligible across the entire scalp due to substantial volume conduction in EEG.
This asymmetry anchors the qualitative reading of the masking grid: increasing $L$ depletes the same-channel temporal context of a masked patch, forcing the predictor to look across channels; increasing $r$ depletes the cross-channel spatial context, forcing it to look across time.
The full characterisation is given in \autoref{app:signal-correlation}.
\autoref{fig:oeb-grid} then asks how this redundancy structure shapes what the model learns: does masking along the temporal axis produce worse representations than masking across the spatially redundant channel axis? Four findings answer this question.

\paragraph[Finding 1]{Finding 1: shared optimum and shared failure modes.}\label{par:finding1}
Both frameworks converge on the same optimal mask, $(L,r) = (2\,\text{s}, 9\,\text{cm})$, and on the same failure modes (\autoref{fig:oeb-grid}\;(A,B)). The redundancy argument predicts both:
at $r=\texttt{"one"}$, the masked patch is trivially predictable from its same-channel temporal neighbors, so the pretext provides too little learning signal;
at $r=\texttt{"all"}$, spatial context is entirely removed, leaving only the weakly predictive temporal axis --- a degenerate regime that heavily impacts JEPA (\nameref{par:finding3}).
Most long temporal blocks ($L \in \{8, 16\}$ at moderate $r$) similarly underperform, consistent with temporal redundancy decaying faster than spatial redundancy.
A notable exception is $L=33$, where the block spans the full window and the mask reduces to masking entire channels: pure cross-channel prediction, hard but tractable given the strong spatial redundancy of EEG.
This regime is not the end of the $L$ continuum but a different pretext task: a fully masked channel can only be inferred from its scalp position, whereas at $L \le 16$ the masked channel remains partly visible and its relation to the neighbouring channels can be read off the visible part.
MAE handles this well (three $L=33$ cells in the ex-aequo cluster); JEPA less so (one cell).

\paragraph[Finding 2]{Finding 2: performance is robust outside the failure modes.}\label{par:finding2}

Once the failure modes are excluded, the choice of $(L, r)$ matters surprisingly
little: $11$ MAE and $9$ JEPA configurations are statistically indistinguishable
from their respective framework's best under the hierarchical bootstrap
(white circles in \autoref{fig:oeb-grid}\;(B)).
The \emph{ex-aequo} cluster spans moderate radii ($r \in \{6, 9, 12\}\,\text{cm}$) and
short lengths ($L \in \{1, 2, 4\}$), with a secondary lobe at $L=33$, consistent
with the asymmetry identified in \nameref{par:finding1}, where full-channel masking is hard
but tractable.
This robustness is not symmetric across frameworks: the within-framework spread
(best minus worst normalised score) is $0.32$ for MAE and $0.77$ for JEPA,
meaning a poor mask choice is $2.4\times$ more costly under JEPA than under MAE.

\paragraph[Finding 3]{Finding 3: bias-inflation collapse of JEPA at $r=$"all".}\label{par:finding3}
\looseness=-1
\autoref{fig:oeb-grid}\;(C) reveals two patterns.
First, the across-dataset normalised score increases with pre-training epochs for nearly all configurations: pre-training leads to downstream performance improvements under most masks.
Second, the $r=\texttt{"all"}$ panel exposes a clear exception that affects only JEPA: the MAE bundle plateaus below the REVE baseline but does not deteriorate, whereas the JEPA bundle descends systematically from epoch~$\sim\!3$ onwards, so more pre-training makes the downstream score worse.
We name this regime \emph{bias-inflation collapse} and discuss its mechanism in \nameref{par:bic-discussion}; a complete analysis is given in \autoref{app:bias-inflation-collapse}.

\paragraph[Finding 4]{Finding 4: REVE-Base-level performance for a fraction of its compute.}\label{par:finding4}
\looseness=-1
At the optimal mask $(L, r) = (2\,\text{s}, 9\,\text{cm})$, both frameworks reach the level of the REVE baseline (\autoref{fig:oeb-best}), while each of our pre-training runs costs around $14$~H100-hours (\autoref{app:compute-cost}), compared to the $260$~A100-hours reported for REVE-Base~\citep{ouahidiREVEFoundationModel2025}.
This efficiency stems from the whole pipeline — our models are smaller ($12.7$M vs.\ $69$M parameters), pre-trained on less data ($4.4$ vs.\ $6$~TB) on a shorter schedule — not from mask geometry alone: under the same pipeline and budget, the failure modes of \nameref{par:finding1} fall well short of this level. Beyond this study, the fact that a competitive model can be pre-trained in $14$~H100-hours lowers the barrier for future work; our open codebase is designed to let researchers rapidly prototype and evaluate new EEG foundation model ideas, and calls for systematic scaling-law studies in this domain.

\section{Discussion and limitations}

\looseness=-1
Under the shared pipeline adopted in this work, the masking strategy is the dominant lever of downstream performance: the within-framework spread ($0.32$ normalised score for MAE, $0.77$ for JEPA) exceeds the cross-framework gap at the optimum, where MAE and JEPA perform comparably.
With well-chosen mask parameters $(L, r) = (2\,\text{s}, 9\,\text{cm})$, both frameworks reach REVE-Base-level downstream performance at a fraction of its pre-training compute.

\paragraph{Why does the mask matter?}\label{par:why-mask-matters}
\looseness=-1
Beyond reporting which masks work, the pattern of failures and successes reveals something deeper: the masking geometry acts as a probe of the signal's redundancy structure, and downstream performance reflects how well the pretext task forces the network to exploit non-trivial statistical dependencies in the data.
At $r=\texttt{"one"}$, the high spatial redundancy of EEG makes the pretext trivially solvable from same-time context alone.
The network never needs to integrate information across times, and the representations it learns carry little structure beyond spatial smoothness.
At $r=\texttt{"all"}$, the extreme opposite, the spatial context is entirely removed: the only surviving signal is the weakly predictive temporal axis, and the pretext degenerates into a task the network solves by ignoring the EEG content altogether.
The successful regimes sit between these extremes, when the mask removes enough spatial redundancy to make the pretext non-trivial, yet leaves enough context for the network to learn genuinely informative representations.
This suggests a general principle for SSL on multichannel biosignals: a good mask should be calibrated to the signal's redundancy structure, targeting the axis along which information is shared but not trivially accessible.

\paragraph[Bias-inflation collapse]{Bias-inflation collapse: a multichannel-specific failure mode of JEPA.}\label{par:bic-discussion}
\looseness=-1
At $r=\texttt{"all"}$, JEPA admits a degenerate solution that MAE structurally cannot: MAE has no teacher network and its target is the per-channel raw input patch, so any failure to reproduce channel-specific signals is penalized at training time; under JEPA, the target is a contextual latent feature produced by the EMA teacher, and the predictor can, in principle, match this target by reproducing a per-channel constant from positional embeddings alone, without ever needing to resolve the EEG content.
This shortcut is unlocked specifically at $r=\texttt{"all"}$: when every channel of the same time-patch is masked together, no same-time spatial neighbor survives in the predictor's context, so the easiest student-teacher agreement collapses onto the position-only lookup table; at smaller $r$, spatial neighbors preserve a non-trivial input-driven path that the optimizer preferentially follows.
We confirm in \autoref{app:bias-inflation-collapse} that this is the regime the optimizer converges to: the encoder learns a trivial channel-position lookup table, the per-channel feature mean inflates by a factor of $2$ to $3$ while the input-driven residual shrinks by a factor of $5$ to $10$, and standard variance-preserving regularizers (VICReg, C-JEPA) do not rescue the regime because they are themselves satisfied by the inflated lookup-table representation.
The collapse is spectrally selective, keeping broadband amplitude but losing the slow structure of the signal (\autoref{app:bic-spectral}), and it persists with $0.5$\,s patches (\autoref{app:ablation-patch-length}), so it is not an artefact of the tokeniser.
The methodological implication is direct: the in-domain collapse metrics computed at training time (variance, EMA divergence, loss thresholds) are insufficient to validate JEPA pre-training in the multichannel setting, and building a reliable in-training detector is itself non-trivial on multi-montage corpora (\autoref{app:bic-detection-challenges}).
Until such a detector exists, a downstream probe is required to expose the failure, and the practical mitigation is simply to avoid the masking regime that triggers it ($r=\texttt{"all"}$) --- consistent with the recommendations of \nameref{par:finding2}. We insist, however, that the deeper concern is not that $r=\texttt{"all"}$ fails, but that the failure is \emph{invisible} to every standard diagnostic.

\paragraph{What we recommend to practitioners.}
\looseness=-1
For practitioners pre-training EEG foundation models, we recommend $(L, r) = (2\,\text{s}, 9\,\text{cm})$ as a robust default under both pretexts; within the ex-aequo cluster ($r \in \{6, 9, 12\}\,\text{cm}$, $L \in \{1, 2, 4\}$) further tuning is unlikely to be productive.
The recommendation holds on two other backbones (\autoref{app:ablation-backbones}) and across mask ratios (\autoref{app:ablation-rho}).
Avoid $r=\texttt{"one"}$ (pretext too easy, and its best $\rho$ depends on the montage; \autoref{app:ablation-rho}) and $r=\texttt{"all"}$ (bias-inflation collapse under JEPA, underperformance under MAE).
If JEPA is the chosen framework, mask choice requires more care than under MAE, whose stability across the grid makes it more forgiving.
Finally, optimise the mask before scaling compute: at a fixed budget, a poor mask costs up to $0.77$ normalised score.

\paragraph{Limitations.}
\looseness=-1
The core grid uses a single backbone and a single pre-training corpus (dominated by clinical EEG and BCI paradigms); transfer to other corpora, hyperparameters, and consumer-grade EEG remains to be established.
The grid fixes the mask ratio at $\rho = 0.55$; $\rho$ and $(L, r)$ interact (\autoref{app:ablation-rho}), and a full joint sweep remains to be done.
One of the $12$ downstream datasets, PhysioNet-MI, is part of our pre-training corpus ($48.5$~h out of $\sim\!34{,}000$~h); the other eleven are unseen.
The REVE-Base checkpoint, our main baseline, additionally saw TUAB and TUEV, and scores above both our $(2\,\text{s}, 9\,\text{cm})$ models on all three, so this exposure does not inflate our results.
More broadly, OpenEEGBench overlaps with the pre-training corpora of other recent EEG foundation models, so collective benchmark overfitting cannot be ruled out.

\newpage
\subsection*{AI use statement}
In this work, we used generative AI tools for one task with required disclosure: implementing methods.
This covers code autocompletion while programming, as well as some larger implementation tasks carried out more autonomously from detailed specifications written by the authors; the architecture of the code base and all its design principles are the authors' own.
We have not used generative AI tools to develop theoretical models or conceptual frameworks, to formulate mathematical claims, to propose or refine hypotheses, to design or provide feedback on the research methodology or experiments, to assist with translation, or to interpret results: the design of all experiments and the interpretation of all results are the authors' own.
Generating synthetic data, proving mathematical claims or writing proofs, cleaning or reformatting datasets, and qualitative or thematic data analysis are not applicable to this work.
Additionally, we used generative AI tools for tasks with recommended disclosure: editing the paper for grammar, spelling and word choice, drafting parts of the paper, searching for relevant literature, and facilitating the running of experiments (monitoring cluster jobs and running smoke tests).
We have reviewed all AI-assisted work, and we take responsibility for the final content of this work, including text, claims, code and other artifacts produced with the aid of generative AI.

\subsection*{Ethics statement}
This work uses only previously collected, publicly available EEG recordings of human participants, released by their original authors: the open-licence subset of the REVE pre-training corpus, whose sources are cited in \autoref{app:datasets}, and the $12$ downstream datasets of OpenEEGBench, described in \autoref{app:downstream-details}.
No new data were recorded and no interaction with human participants took place for this study.
We restricted pre-training to the open-licence part of the corpus so that the resulting checkpoints and code can be redistributed.
Beyond the general dual-use concerns of brain-signal decoding, such as inferring cognitive or clinical states without informed consent, which apply to the field as a whole, we do not foresee harmful applications specific to this work.

\subsection*{Reproducibility statement}
All $58$ models share the single recipe of \autoref{sec:method}, so reproducing the study amounts to reproducing one pipeline under $29$ mask configurations and two objectives.
The code for pre-training and downstream evaluation, together with the cached results of the main grid, is available at the link given in the introduction.
The backbone is specified in \autoref{app:reve-small-arch}, the training hyperparameters in \autoref{app:training-hyperparams}, the compute cost in \autoref{app:compute-cost}, and the two objectives in Appendices~\ref{app:mae-details} and~\ref{app:jepa-details}.
The mask sampler is fully determined by \eqref{eq:K} and Algorithm~\ref{alg:mask-maker} (\autoref{app:mask-algorithm}), and its realised mask ratio is checked in \autoref{app:mask-calibration}.
The pre-training corpus is listed in \autoref{app:datasets}; the downstream datasets, splits, probe and score normalisation are given in \autoref{app:downstream-details}, and the statistical testing procedure in \autoref{sub:downstream}.

\ificlrfinal
\subsubsection*{Acknowledgments}
  \paragraph{Author contributions.}
  PG designed and ran the experiments, performed the analysis, and wrote the paper.
  BA ran some of the experiments and reviewed the manuscript.
  YEO contributed to the writing and provided the pre-training corpus: a 4.4\,TB dataset aggregating and pre-processing publicly available EEG datasets, originally assembled for prior work~\citep{ouahidiREVEFoundationModel2025} (no new data were recorded for the present study; individual sources are listed in \autoref{app:datasets}).
  AD reviewed the manuscript.
  TM and MT contributed to the experimental design and to the writing.

  \paragraph{Funding and competing interests.}
  This work used the Dutch national e-infrastructure with the support of the SURF Cooperative using grant no.~EINF-14502.\\
  TM is supported by the ANR EBUL (ANR-23-CE23-0001).
  Additionally, we acknowledge Amitava Majumdar for his leadership of the Neuroscience Gateway (NSG) project at the San Diego Supercomputer Center, which provides community access to high-performance computing resources for large-scale neuroscience research.
  Funding source for NSG GPU allocation: NAIRR250045.

  \paragraph{Additional acknowledgments.}
  PG thanks Göksenin Yüksel for the insightful discussions on the JEPA framework.
\fi

\bibliography{references,references_datasets,references_datasets_downstream}
\bibliographystyle{iclr2027_conference}

\appendix

\section{Frameworks details}\label{app:frameworks-details}\label{app:frameworks-shared-details}

\subsection{Detailed REVE-Small architecture}\label{app:reve-small-arch}

The REVE-Small backbone follows the REVE-Small architecture of
\citet{ouahidiREVEFoundationModel2025} with minor simplifications
that we keep across all pre-training configurations.
The pipeline goes as follows: a shared linear patch embedding turns each EEG channel into a sequence
of overlapping temporal patches of length $200$ samples ($1\,\text{s}$
at $200\,\text{Hz}$) with $20$-sample ($0.1\,\text{s}$) overlap, each
linearly projected to $d=512$ dimensions; the same embedding is reused
by the student encoder, the EMA teacher (JEPA only) and the predictor.
A non-learnable additive $4$D sinusoidal positional encoder
encodes the channel position $(x,y,z)$ on $d_\text{spat}=384$
dimensions ($128$ per coordinate) and the patch index on
$d_\text{time}=128$ dimensions, computed at the post-embedding sampling
rate $\nu_\text{feat}=200/180=10/9\,\text{Hz}$;
its parameters are buffers, not weights, so it does not appear in the
parameter count below. The contextual encoder is a $4$-layer
transformer with $8$ heads, $d_\text{model}=512$,
GEGLU~\citep{shazeerGLUVariantsImprove2020} feed-forward blocks at the
LLaMA $8/3$ expansion ratio ($d_\text{ff}=\lfloor 8\cdot 512/3\rfloor =
1365$), pre-norm RMSNorm~\citep{zhangRootMeanSquare2019}, no bias on
linear layers, and dropout $0$.
The MAE decoder and the JEPA predictor share a single class:
a $2$-layer transformer decoder with the same width, head count, FFN
geometry, and norm; each layer applies pre-norm self-attention over the
mask tokens, cross-attention to the encoder output (memory), and a
GEGLU feed-forward block.
The decoder query is a single learned $d$-dimensional
mask token (no separate positional embedding for the mask
target: the additive PE from the encoder is reused). A final
$d_\text{model}\!\to\!p$ linear layer maps each output token to the
prediction target, with $p=200$ for MAE (raw patch in input space) and
$p=512$ for JEPA (latent embedding of the same dimension as the
encoder).

Table~\ref{tab:reve-small-params} reports the per-component parameter
count, derived directly from the layer specifications above and from the
\texttt{torch.nn} primitives we use (notably: in our GEGLU FFN, the
  input projection maps $d_\text{model}\!\to\!2\,d_\text{ff}$ so that
  the gate-and-multiply activation can halve its output back to
  $d_\text{ff}$; RMSNorm contributes one learnable scale per channel;
\texttt{bias=False} on every linear layer).
The MAE student totals $\sim 21.2$M parameters and the JEPA student
$\sim 21.3$M parameters; the JEPA EMA teacher is a frozen copy of the
encoder only ($\sim 12.6$M parameters held in fp32 memory but excluded
from the optimiser, see Appendix~\ref{app:jepa-details}); the patch
embedding is shared with the student rather than duplicated.

\begin{table}[!b]
  \centering
  \caption{Per-component parameter count of the REVE-Small backbone, derived
    from the configuration files. ``$\dagger$'' marks components that exist
    only for one of the two frameworks; ``$\ddagger$'' marks the JEPA-only
    EMA teacher, which is held in memory but not optimised. Numbers assume
    \texttt{bias=False} on every linear layer and the GEGLU expansion
  described in the prose.}
  \label{tab:reve-small-params}
  \footnotesize
  \begin{tabular}{lr}
    \toprule
    Component & \# params \\
    \midrule
    Linear patch embedding ($200\!\to\!512$, with bias) & 102{,}912 \\ %
    Encoder layer (single, $\times 4$): & 3{,}146{,}240 \\ %
    \quad\,\,Self-attention $W_{QKV}$, $W_O$ & 1{,}048{,}576 \\
    \quad\,\,GEGLU FFN (gate + value, output projections) & 2{,}096{,}640 \\
    \quad\,\,$2\times$ RMSNorm scales & 1{,}024 \\
    Encoder transformer total (4 layers) & 12{,}584{,}960 \\
    Decoder layer (single, $\times 2$): & 4{,}195{,}328 \\ %
    \quad\,\,Self-attention $W_{QKV}$, $W_O$ & 1{,}048{,}576 \\
    \quad\,\,Cross-attention $W_{QKV}$, $W_O$ & 1{,}048{,}576 \\
    \quad\,\,GEGLU FFN (gate + value, output projections) & 2{,}096{,}640 \\
    \quad\,\,$3\times$ RMSNorm scales & 1{,}536 \\
    Decoder transformer total (2 layers) & 8{,}390{,}656 \\
    Learned mask token ($d=512$) & 512 \\ %
    Output projection $\textsuperscript{$\dagger$}$, MAE ($512\!\to\!200$) & 102{,}400 \\ %
    Output projection $\textsuperscript{$\dagger$}$, JEPA ($512\!\to\!512$) & 262{,}144 \\ %
    \midrule
    \textbf{MAE student total} & \textbf{21{,}181{,}440} \\
    \textbf{JEPA student total} & \textbf{21{,}341{,}184} \\
    JEPA EMA teacher$\textsuperscript{$\ddagger$}$ (encoder only; patch embedding shared with student) & 12{,}584{,}960 \\ %
    \bottomrule
  \end{tabular}
\end{table}

\subsection{Training hyperparameters}\label{app:training-hyperparams}

The same optimizer, learning-rate schedule, batch size, precision, and gradient clipping are used for every one of the $58$ pre-training configurations of the core grid; only the framework loss and the masking strategy change between configurations. The values in
Table~\ref{tab:training-hyperparams} are copied verbatim from the launch
scripts and traced to the corresponding default in the codebase when the
launch script does not override them.

\begin{table}[!h]
  \centering
  \caption{Training hyperparameters shared by all $58$ pre-training
  configurations of the core grid.}
  \label{tab:training-hyperparams}
  \footnotesize
  \begin{tabular}{ll}
    \toprule
    Hyperparameter & Value \\
    \midrule
    Optimizer & AdamW \\
    \quad $(\beta_1,\beta_2)$ & $(0.9,\,0.999)$ (PyTorch default) \\
    \quad weight decay & $0.01$ \\
    Peak learning rate & $2.4\!\times\!10^{-4}$ \\
    Final learning rate $\eta_\text{min}$ & $10^{-6}$ \\
    Warmup schedule & linear, factor $10^{-6}\!\to\!1$ \\
    Warmup steps & $3{,}080$ ($=1$ epoch) \\
    LR schedule (after warmup) & cosine, $T_\text{max}=T_0-\text{warmup}$ \\
    $T_0$ & $30{,}800$ ($=$ total optimizer steps) \\
    Total epochs & $10$ \\
    Steps per epoch & $3{,}080$ ($=\lceil 3{,}695{,}009 / 1200\rceil$) \\
    Total optimizer steps & $30{,}800$ \\
    Per-GPU batch size & $600$ \\
    Number of GPUs (DDP) & $2$ ($1$ \ificlrfinal Snellius \fi H100 node) \\
    Effective batch size & $1{,}200$ windows / step \\
    Mixed-precision dtype & \texttt{bf16-mixed} \\
    Gradient clipping & $\|g\|_2 \le 3.0$ \\
    DDP strategy & \texttt{ddp\_find\_unused\_parameters\_true} \\
    Random seed & $42$ \\
    \bottomrule
  \end{tabular}
\end{table}

The cosine schedule is wrapped in a \texttt{SequentialLR} that switches
from a $1$-epoch linear warm-up to
\texttt{torch.optim.lr\_scheduler.CosineAnnealingLR} with
$T_\text{max}=T_0-\text{warmup\_steps}=27{,}720$ at the warm-up
milestone. %
We deliberately set $T_0$ \emph{exactly} equal to the total optimizer
step count: a smaller $T_0$ would let \texttt{CosineAnnealingLR} pass
its minimum and the learning rate would climb back up over the last
steps, which we observed to wreck the final weights in earlier sweeps.

\subsection{Pre-training compute cost}\label{app:compute-cost}

All $58$ grid runs share the same $10$-epoch budget and the same hardware, two H100 GPUs on a single node (\autoref{tab:training-hyperparams}).
We report the cost of each run in H100-hours, \emph{i.e.}\ its measured wall-clock duration multiplied by the number of GPUs.
\autoref{tab:compute-cost} summarises these costs per framework: JEPA runs cost about $2$~H100-hours more than MAE runs, and the spread within each framework is small.
For reference, \citet{ouahidiREVEFoundationModel2025} report an estimated $260$~A100-hours for the original REVE-Base pre-training.
On our cluster, an H100-hour is billed $1.5\times$ an A100-hour; at this rate, the original REVE-Base pre-training costs about $12\times$ one of our runs.
That comparison is not scale-matched, since REVE-Base has $69$M parameters and was trained on the full $6$~TB corpus whereas our models have $12.7$M parameters and use the $4.4$~TB open subset, and the convergence trajectory of REVE-Base is not public; the gap in compute therefore cannot be attributed to the mask geometry alone.

\begin{table}[!h]
  \centering
  \caption{Pre-training cost of the $58$ grid runs, in H100-hours (two GPUs per run; $n$ runs per group).}
  \label{tab:compute-cost}
  \footnotesize
  \begin{tabular}{lrrrrr}
    \toprule
    Framework & $n$ & Mean & Median & SD & Min--Max \\
    \midrule
    JEPA    & $29$ & $14.91$ & $15.28$ & $0.78$ & $14.00$--$16.29$ \\
    MAE     & $29$ & $12.70$ & $12.65$ & $0.25$ & $12.62$--$13.98$ \\
    Overall & $58$ & $13.81$ & $13.99$ & $1.26$ & $12.62$--$16.29$ \\
    \bottomrule
  \end{tabular}
\end{table}

\subsection{Mask sampling algorithm}\label{app:mask-algorithm}

For each pre-training configuration the masking strategy is fully
specified by the pair $(L, r)$ where $L\in\mathbb{N}_{>0}$ is the
temporal block length in patch units and $r\in[0,\infty]$ is the spatial
cap radius, together with the target mask ratio
$\rho^{\star}=0.55$ shared across all $58$ configurations of the core grid.
The mask sampler computes the number of independently sampled
blocks $K$ from the inclusion-exclusion formula and then samples the
$K$ blocks; Algorithm~\ref{alg:mask-maker} states the procedure as it
appears in the codebase.

\begin{algorithm}[t]
  \small
  \caption{Spatio-temporal block-mask sampling.
    Inputs are the $(L,r)$ axis values, the channel positions
    $\{p_c\}\subset\mathbb{R}^3$, the optional channel-padding mask, the
    target mask ratio $\rho^{\star}$, the scalp surface
    $A_\text{scalp}=4\pi(0.1)^{2}\cdot 3/4$ ($3/4$ of a $10\,\text{cm}$
  sphere), and the patch-grid size $(N_c, N_t)$.}
  \label{alg:mask-maker}
  \begin{algorithmic}[1]
    \Function{$f_\text{sp}$}{$r$, $N_c$}
    \If{$r = 0$} \State \Return $1/N_c$ \Comment{single-channel block (special case)}
    \ElsIf{$r = \infty$} \State \Return $1$ \Comment{full-spatial block (special case)}
    \Else \State \Return $\pi r^{2}/A_\text{scalp}$ \Comment{small-cap surface fraction}
    \EndIf
    \EndFunction
    \State $f \gets f_\text{sp}(r, N_c)\cdot L/N_t$ \Comment{per-block cell coverage}
    \If{$f \ge 1$} \State $K\gets 1$
    \Else \State $K \gets \max\left(1,\,\left\lceil\frac{\log(1-\rho^{\star})}{\log(1-f)}\right\rceil\right)$ \Comment{inclusion-exclusion, see Eq.~\eqref{eq:K}}
    \EndIf
    \State $M \gets$ all-zeros boolean tensor of shape $(N_c, N_t)$
    \For{$i = 1,\ldots,K$}
    \State $c_i \gets$ sample uniformly from the non-padding channels
    \State $t_i \gets$ sample uniformly from $\{1,\ldots,\max(1,N_t-L+1)\}$
    \For{$c = 1,\ldots,N_c$}
    \If{$r=0$} \State $\text{in}\gets[c=c_i]$ \Comment{distance $\le 0$ matches only $c_i$}
    \ElsIf{$\|p_c-p_{c_i}\|_2\le r$} \State $\text{in}\gets\text{True}$
    \Else \State $\text{in}\gets\text{False}$
    \EndIf
    \If{$\text{in}$}
    \For{$t=t_i,\ldots,t_i+L-1$}
    \State $M_{c,t}\gets\text{True}$
    \EndFor
    \EndIf
    \EndFor
    \EndFor
    \State \Return $M$ \Comment{$M_{c,t}=1$ iff cell $(c,t)$ is masked}
  \end{algorithmic}
\end{algorithm}

\subsection{Empirical validation of the block count $K$ and mask fraction $\rho^{\star}$}\label{app:mask-calibration}

We sanity-check the inclusion-exclusion formula for $K$ (\autoref{eq:K}) by counting, in simulation, the realised cell-level mask fraction it produces under our pre-training
sampling pipeline. For each $(L, r)$ pair we draw $1{,}000$
batches of $64$ examples and aggregate over the resulting $64{,}000$
samples.

\begin{table}[!t]
  \centering
  \caption{
    Number of blocks $K$ sampled for an input $x$, for each combination radius-length $(r, L)$, derived from the inclusion-exclusion formula \eqref{eq:K} with the mask ratio $\rho^{\star}=0.55$.
  }
  \label{tab:K-grid}
  \small
  \begin{tabular}{lcccccc}
  \toprule
  & \multicolumn{6}{c}{Temporal block length $L$ (patches)} \\
  \cmidrule(lr){2-7}
  Spatial radius $r$ & $1$ & $2$ & $4$ & $8$ & $16$ & $33$ \\
  \midrule
  $0.0$ ($r=$\texttt{"one"}) & 843 & 422 & 211 & 106 & 53 & 26 \\
  $0.06\,\text{m}$ & 220 & 110 & 55 & 28 & 14 & 7 \\
  $0.09\,\text{m}$ & 98 & 49 & 24 & 12 & 6 & 3 \\
  $0.12\,\text{m}$ & 55 & 28 & 14 & 7 & 4 & 2 \\
  $\infty$ ($r=$\texttt{"all"}) & 26 & 13 & 7 & 3 & 2 & $\boldsymbol{-}$ \\
  \bottomrule
\end{tabular}

\end{table}

The number of blocks $K$ given by \eqref{eq:K} for each of the $29$ cells is listed in \autoref{tab:K-grid}; it spans more than two orders of magnitude, from $K=2$ for the largest blocks to $K=843$ for single-patch blocks.

Table~\ref{tab:mask-coverage} reports two empirical quantities per
$(L, r)$ cell: the per-block channel coverage (\emph{ch/sphere}, the
  mean and standard deviation, over individual blocks, of the number of
  non-padding channels falling inside the spherical cap of one randomly
drawn block) and the realised cell-level mask fraction (\emph{total
  mask}, the mean and standard deviation, over the $64{,}000$ samples,
  of the fraction of $(c,t)$ tokens flagged as masked after the $K$
blocks are merged). The \emph{ch/sphere} value is independent of $L$
for fixed $r$, as expected. The realised mask fraction matches the
$\rho^{\star}=55\,\%$ target within a few points across most cells,
with the largest deviations observed in the extreme regimes $r=\texttt{"all"}$ and $L=33$.

\begin{table}[!t]
  \centering
  \caption{Empirical per-block channel coverage (\emph{ch/sphere}) and
    realised total mask fraction (\emph{total mask}) for every
    $(L, r)$ cell, with $K$ as in \autoref{tab:K-grid}; mean
    $\pm$~std over $1{,}000\times 64=64{,}000$ samples. Target mask
  fraction is $\rho^{\star}=55\,\%$.}
  \label{tab:mask-coverage}
  \footnotesize
  \resizebox{\linewidth}{!}{%
    \begin{tabular}{llcccccc}
  \toprule
  & & \multicolumn{6}{c}{Temporal block length $L$ (patches)} \\
  \cmidrule(lr){3-8}
  Spatial radius $r$ & metric & $1$ & $2$ & $4$ & $8$ & $16$ & $33$ \\
  \midrule
  \multirow{2}{*}{$0.0$ ($r=$\texttt{"one"})} & ch / sphere & $1.00 \pm 0.00$ & $1.00 \pm 0.00$ & $1.00 \pm 0.00$ & $1.00 \pm 0.00$ & $1.00 \pm 0.00$ & $1.00 \pm 0.00$ \\
   & tot \% & $55.0 \pm 1.1$ & $54.8 \pm 1.3$ & $54.2 \pm 1.8$ & $53.1 \pm 2.5$ & $50.9 \pm 3.4$ & $55.8 \pm 6.3$ \\
  \cmidrule(lr){1-8}
  \multirow{2}{*}{$0.06\,\text{m}$} & ch / sphere & $3.08 \pm 1.60$ & $3.08 \pm 1.60$ & $3.09 \pm 1.61$ & $3.08 \pm 1.60$ & $3.09 \pm 1.60$ & $3.09 \pm 1.60$ \\
   & tot \% & $56.2 \pm 4.1$ & $56.0 \pm 4.2$ & $55.5 \pm 4.5$ & $54.7 \pm 5.1$ & $53.0 \pm 5.8$ & $60.3 \pm 10.0$ \\
  \cmidrule(lr){1-8}
  \multirow{2}{*}{$0.09\,\text{m}$} & ch / sphere & $6.09 \pm 2.54$ & $6.08 \pm 2.52$ & $6.10 \pm 2.54$ & $6.10 \pm 2.54$ & $6.09 \pm 2.53$ & $6.10 \pm 2.54$ \\
   & tot \% & $53.5 \pm 3.4$ & $53.4 \pm 3.6$ & $52.3 \pm 4.2$ & $51.6 \pm 5.2$ & $50.8 \pm 6.3$ & $59.0 \pm 11.0$ \\
  \cmidrule(lr){1-8}
  \multirow{2}{*}{$0.12\,\text{m}$} & ch / sphere & $10.53 \pm 3.62$ & $10.53 \pm 3.63$ & $10.52 \pm 3.62$ & $10.54 \pm 3.63$ & $10.51 \pm 3.61$ & $10.52 \pm 3.63$ \\
   & tot \% & $53.2 \pm 3.3$ & $53.8 \pm 3.8$ & $53.6 \pm 4.7$ & $53.4 \pm 6.2$ & $57.5 \pm 7.5$ & $70.0 \pm 13.3$ \\
  \cmidrule(lr){1-8}
  \multirow{2}{*}{$\infty$ ($r=$\texttt{"all"})} & ch / sphere & $23.01 \pm 6.01$ & $23.01 \pm 6.01$ & $23.05 \pm 6.03$ & $23.04 \pm 6.02$ & $23.02 \pm 6.02$ & $\boldsymbol{-}$ \\
   & tot \% & $55.1 \pm 5.1$ & $55.4 \pm 6.1$ & $58.5 \pm 8.2$ & $54.6 \pm 10.7$ & $66.6 \pm 12.8$ & $\boldsymbol{-}$ \\
  \bottomrule
\end{tabular}

  }
\end{table}

\begin{figure}[t]
  \centering
  \includegraphics[width=\linewidth]{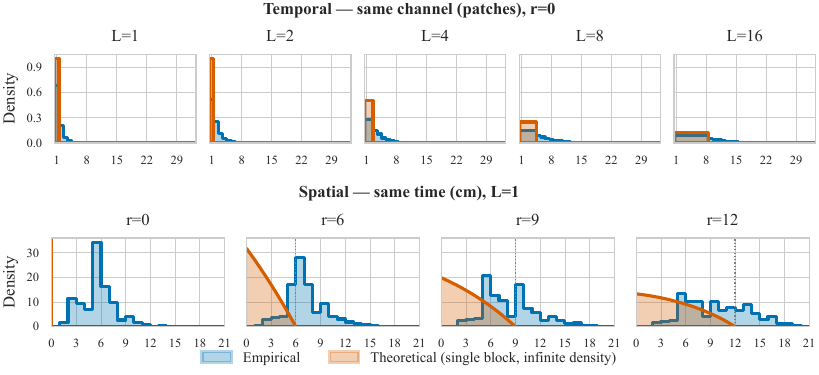}
  \caption{Distance from each masked token to its nearest unmasked token, empirical (blue) vs.\ single-block infinite-density theoretical (orange).
    \emph{Top row:} temporal distance (in patches) on the same channel, sweeping $L$ at $r=\texttt{"one"}$.
    \emph{Bottom row:} spatial distance (in centimeters) at the same time, sweeping $r$ at $L=1$.
    Empirical distributions are aggregated over $1{,}000$ batches of $64$ samples drawn from the REVE channel layouts.
  }
  \label{fig:mask-token-distances}
\end{figure}

\paragraph{Mask geometry: empirical vs.\ theoretical token distances.}
\autoref{fig:mask-token-distances} reports the distance from each masked token to its nearest unmasked token on the same channel (top row, varying $L$ at $r=\texttt{"one"}$) and at the same time on a different channel (bottom row, varying $r$ at $L=1$), comparing the empirically realised distribution under our pre-training pipeline to the analytical single-block, infinite-density reference.
The empirical and theoretical distributions track each other closely across the entire $(L, r)$ grid, confirming that the inclusion-exclusion formula of \autoref{eq:K} produces the intended geometry on the actual REVE channel layouts: longer $L$ pushes a masked patch further from the nearest unmasked patch \emph{on the same channel} (so the network must look across channels to fill it in), and larger $r$ pushes it further from the nearest unmasked patch \emph{at the same time} (so the network must look across time).
The cell $r=\texttt{"all"}$ is degenerate by construction: every channel is masked at masked time-points, so the spatial distance is undefined and only the temporal axis carries information about the masked tokens.

\subsection{MAE details}\label{app:mae-details}

The MAE framework reconstructs the \emph{raw
input patches} at masked positions.
Concretely, the shared linear patch embedding returns both the
$d$-dimensional embeddings (used as the encoder's local features) and
the unfolded raw patches $\mathbf{P}\in\mathbb{R}^{B\times C\times
T\times p}$ with $p=200$ samples per patch; the predictor
output $\hat{\mathbf{P}}\in\mathbb{R}^{B\times C\times T\times p}$ comes
from the same $2$-layer transformer decoder as in JEPA but with output
dimension $p=200$ (the input patch length) rather than $d=512$.
The objective is a masked MSE
\[
  \mathcal{L}_\text{MAE}
  =
  \frac{1}{|\mathcal{M}|\,p}
  \sum_{(c,t)\in\mathcal{M}}
  \|\hat{\mathbf{P}}_{c,t}-\mathbf{P}_{c,t}\|_{2}^{2},
\]
identical in form to the JEPA loss but with raw-patch targets; the
target tensor is detached and zero-filled outside the target mask before
the elementwise MSE.
No per-patch target normalisation is applied at the loss level: the input
window has already been scaled by the per-window robust scaler
before
patching, and we found in preliminary runs that the additional
per-patch normalisation used in some computer-vision MAE pipelines was
unnecessary at the REVE-Small scale.
The variance and covariance regularisers exposed by the framework are
unused in the $58$ configurations of the core grid (their loss weights are set to $0$),
so the only term optimised is $\mathcal{L}_\text{MAE}$, preserving
strict parity with the JEPA-no-reg configuration on the ``no auxiliary
loss'' axis.

\subsection{JEPA details}\label{app:jepa-details}

The JEPA framework regresses the predictor's
output at masked positions onto the contextual features of an
exponential-moving-average (EMA) teacher under a masked $L_{2}$ loss;
in our \emph{no-reg} configuration the variance and covariance
regularisers of
\citet{bardesVICRegVarianceInvarianceCovarianceRegularization2021} are
explicitly disabled by setting their weights to $0$, so the only term
optimised is the main prediction loss
$
\mathcal{L}_\text{JEPA}
=
\frac{1}{|\mathcal{M}|\,d}
\sum_{(c,t)\in\mathcal{M}}\|\hat{z}_{c,t}-\bar{z}_{c,t}\|_{2}^{2},
$ where $\mathcal{M}=\{(c,t)\,:\,M_{c,t}=1 \text{ and } (c,t) \text{ is not padding}\}$,
$\hat{z}_{c,t}\in\mathbb{R}^{d}$ is the predictor output, and
$\bar{z}_{c,t}\in\mathbb{R}^{d}$ is the EMA-teacher contextual feature
at the same position; the loss weights at masked, non-padded cells
contribute equally regardless of the per-block density.
The EMA teacher is built once at framework construction time as a
\texttt{deepcopy} of the encoder reduced to its maskless form
— it owns its own copy of
the additive positional encoder and the $4$-layer transformer but holds
no masker, and the patch embedding is shared with the student (its
output is cached during the student's forward and reused as the
teacher's input) rather than duplicated. The teacher is kept in
\texttt{eval()} mode throughout training; it receives the
\emph{unmasked} input, so its forward pass costs an additional
$\sim\!12.6$M-parameter encoder evaluation per step
(Table~\ref{tab:reve-small-params}).
Targets are taken from the last encoder layer only — no multi-layer
averaging of teacher features.

The EMA decay schedule interpolates linearly from
$\tau_\text{start}=0.999$ at step $0$ to $\tau_\text{end}=1.0$ at
step $30{,}000$, after which it is held at $\tau_\text{end}=1.0$ (i.e.\
the teacher is frozen for the last $\sim\!2.6\,\%$ of training):
\[
  \tau(s) =
  \begin{cases}
    \tau_\text{end} - (\tau_\text{end}-\tau_\text{start})\,
    \Bigl(1 - \dfrac{s}{S_\text{anneal}}\Bigr),
    & s < S_\text{anneal} = 30{,}000, \\[1mm]
    \tau_\text{end} = 1.0, & s \ge S_\text{anneal},
  \end{cases}
\]
with $\tau_\text{start}=0.999$ and $\tau_\text{end}=1.0$, where
$\tau$ is the teacher's keep weight (so the per-step update on every
  trainable teacher parameter is
  $\theta_\text{T}\!\leftarrow\!\tau\,\theta_\text{T}+(1-\tau)\,\theta_\text{S}$,
  performed in \texttt{fp32} regardless of the training precision; non-trainable
teacher buffers are copied directly from the student each step).
The teacher update is performed \emph{per training step}, immediately
after the optimiser step; it is not called during validation or test.

\paragraph{Collapse diagnostics for JEPA.}
We monitor JEPA runs for representational collapse throughout
pre-training, but do not add any corrective loss.
The three failure modes we guard against are documented in the
self-supervised learning literature: dimensional/variance collapse,
addressed by the variance term of VICReg
\citep{bardesVICRegVarianceInvarianceCovarianceRegularization2021};
trivial-constant ``entire collapse'', which
\citet{moConnectingJointEmbeddingPredictive2024} show is not
reliably prevented by the EMA teacher of I-JEPA
\citep{assranSelfSupervisedLearningImages2023a} alone and
motivates their addition of VICReg-style regularisation; and the
post-collapse loss divergence that follows once the encoder outputs
degenerate.
A Lightning callback halts the run
as soon as any of three detection conditions fires.
First, the embedding standard deviation $\sigma_z$ logged by the
framework at every optimiser step, the actual standard deviation of
the encoder outputs across the batch, must remain above
$\epsilon = 10^{-2}$; if it falls below $\epsilon$ for $200$
consecutive steps the run is flagged for variance collapse.  This
threshold is well below the unit-variance target $\gamma = 1$ used by
the VICReg variance hinge
\citep{bardesVICRegVarianceInvarianceCovarianceRegularization2021},
so the rule fires only when the encoder has effectively lost its
representational variance.
Second, the main prediction loss exceeds $10^{4}$, indicating
post-collapse divergence of the kind reported by
\citet{moConnectingJointEmbeddingPredictive2024} when the EMA teacher
fails to stabilise training.
Third, the main prediction loss drops below $10^{-3}$ within the
first $500$ optimiser steps, which is the signature of the predictor
latching onto a trivial constant target -- the ``entire collapse''
mode of \citet{moConnectingJointEmbeddingPredictive2024}.
Across all JEPA configurations reported in this paper, none of these
conditions was ever triggered.

\section{Pre-training datasets}\label{app:datasets}

The open subset of the REVE pre-training corpus comprises the following datasets:
\citep{%
  ouahidi2025reve,%
  openneuro_ds004706_v1_0_0,%
  openneuro_ds004582_v1_0_0,%
  openneuro_ds004356_v2_2_1,%
  openneuro_ds004817_v1_0_1,%
  openneuro_ds005189_v1_0_1,%
  openneuro_ds003887_v1_2_3,%
  openneuro_ds004043_v1_1_0,%
  openneuro_ds003885_v1_0_8,%
  openneuro_ds004357_v1_0_1,%
  openneuro_ds003825_v1_2_0,%
  openneuro_ds004816_v1_0_1,%
  openneuro_ds004840_v1_0_1,%
  openneuro_ds005262_v1_0_1,%
  openneuro_ds004477_v1_0_2,%
  openneuro_ds005273_v1_0_0,%
  openneuro_ds004561_v1_0_0,%
  openneuro_ds004951_v1_0_0,%
  openneuro_ds004324_v1_0_0,%
  openneuro_ds005095_v1_0_1,%
  openneuro_ds001787_v1_1_1,%
  openneuro_ds003690_v1_0_0,%
  openneuro_ds004603_v1_1_0,%
  openneuro_ds003969_v1_0_0,%
  openneuro_ds004147_v1_0_2,%
  openneuro_ds003004_v1_1_1,%
  openneuro_ds002721_v1_0_3,%
  openneuro_ds004152_v1_1_2,%
  openneuro_ds005089_v1_0_1,%
  openneuro_ds004262_v1_0_0,%
  openneuro_ds004264_v1_1_0,%
  openneuro_ds004315_v1_0_0,%
  openneuro_ds004408_v1_0_8,%
  openneuro_ds005121_v1_0_2,%
  openneuro_ds003775_v1_2_1,%
  openneuro_ds004572_v1_3_2,%
  openneuro_ds002778_v1_0_5,%
  openneuro_ds003846_v2_0_2,%
  openneuro_ds004279_v1_1_2,%
  openneuro_ds004148_v1_0_1,%
  openneuro_ds004902_v1_0_8,%
  openneuro_ds002680_v1_2_0,%
  openneuro_ds004284_v1_0_0,%
  openneuro_ds004395_v2_0_0,%
  openneuro_ds005505_v1_0_1,%
  openneuro_ds005506_v1_0_1,%
  openneuro_ds005507_v1_0_1,%
  openneuro_ds005509_v1_0_1,%
  openneuro_ds005510_v1_0_1,%
  openneuro_ds005511_v1_0_1,%
  openneuro_ds005512_v1_0_1,%
  openneuro_ds005514_v1_0_1,%
  openneuro_ds005508_v1_0_1,%
  openneuro_ds005697_v1_0_2,%
  openneuro_ds005620_v1_0_0,%
  openneuro_ds005594_v1_0_3,%
  openneuro_ds005586_v2_0_0,%
  barachant2012commande,%
  cho2017eeg,%
  lee2019eeg,%
  liu2024eeg,%
  ofner2017upper,%
  schalk2004bci2000,%
  goldberger2000physiobank,%
  yi2014evaluation,%
  zhou2016fully,%
  schirrmeister2017deep,%
  kalunga2015using,%
  lee2019eegssvep,%
  korczowski2019brain2014a,%
  korczowski2019brain2014b,%
  korczowski2019brain2015a,%
  korczowski2019brain2015b,%
  sosulski2019spatial,%
  lee2019eeg_erp,%
  icare%
}.

\section{Downstream evaluation details} \label{app:downstream-details}

We evaluate every model on the $12$ OpenEEGBench datasets~\citep{guetschelOpenEEGBench2026}.
Foundation-model checkpoints are evaluated under a closed-form ridge linear probe on frozen features (\autoref{app:downstream-fm}); the three non-foundation baselines (EEGNet, EEGConformer, ShallowFBCSPNet) are trained from scratch (\autoref{app:downstream-non-fm}).

\subsection{Downstream datasets}
\label{app:downstream-datasets}

\paragraph{Datasets.}
The 12 OpenEEGBench downstream datasets are
Arithmetic~\citep{zymaElectroencephalogramsMentalArithmetic2019,goldberger2000physiobank},
BCIC-IV-2a~\citep{brunner2008bcic2a},
BCIC2020-3~\citep{jeong2020InternationalBrain2022a},
PhysioNet-MI~\citep{schalk2004bci2000,goldberger2000physiobank},
CHB-MIT~\citep{shoeb2009chbmit,goldberger2000physiobank},
FACED~\citep{chen2023faced},
ISRUC-Sleep~\citep{khalighiISRUCSleepComprehensivePublic2016},
MDD~\citep{mumtaz2016mdd},
SEED-V~\citep{liu2022seedv},
SEED-VIG~\citep{zheng2017seedvig},
TUAB~\citep{obeid2016tuh},
and TUEV~\citep{obeid2016tuh}.

\paragraph{Per-dataset class distributions.}
\label{app:oeb-class-distributions}
\autoref{tab:oeb-class-distributions} reports the per-class window counts and percentages for each of the 12 OpenEEGBench downstream datasets.
\begin{table}[t]
  \centering
  \caption{Per-class window counts and balance for each of the 12 OpenEEGBench downstream datasets, computed from the per-recording metadata of the Hugging Face Zarr mirror.}
  \label{tab:oeb-class-distributions}
  \footnotesize
  \begin{tabular}{lrcl}
    \toprule
    Dataset & $n_\mathrm{windows}$ & classes & Distribution (\%) \\
    \midrule
    \texttt{arithmetic\_zyma2019} & $1{,}707$   & $2$ & $25.3 \,/\, 74.7$ \\
    \texttt{bcic2020-3}           & $6{,}000$   & $5$ & $20 \,/\, 20 \,/\, 20 \,/\, 20 \,/\, 20$ \\
    \texttt{bcic2a}               & $5{,}184$   & $4$ & $25 \,/\, 25 \,/\, 25 \,/\, 25$ \\
    \texttt{chbmit}               & $37{,}046$  & $2$ & $\mathbf{2.5 \,/\, 97.5}$ \\
    \texttt{faced}                & $19{,}217$  & $8$ & $13 \,/\, 14 \,/\, 13 \,/\, 12 \,/\, \text{---} \,/\, 12 \,/\, 17 \,/\, 7.7 \,/\, 12$ \\
    \texttt{isruc-sleep}          & $98{,}480$  & $5$ & $21 \,/\, 14 \,/\, 33 \,/\, 19 \,/\, 14$ \\
    \texttt{mdd\_mumtaz2016}      & $14{,}743$  & $2$ & $47 \,/\, 53$ \\
    \texttt{physionet}            & $9{,}688$   & $4$ & $25 \,/\, 25 \,/\, 25 \,/\, 25$ \\
    \texttt{seed-v}               & $117{,}744$ & $5$ & $17 \,/\, 26 \,/\, 20 \,/\, 17 \,/\, 20$ \\
    \texttt{seed-vig}             & $20{,}355$  & \textbf{regression} & range $[0.022,\, 1.0]$, mean $0.467$ \\
    \texttt{tuab}                 & $42{,}136$  & $2$ & $49.7 \,/\, 50.3$ \\
    \texttt{tuev}                 & $3{,}330$   & $6$ & $\mathbf{2.4} \,/\, 19 \,/\, 23 \,/\, 8.1 \,/\, 16 \,/\, \mathbf{31}$ \\
    \bottomrule
  \end{tabular}
\end{table}

\paragraph{Signal scaling.}
For every foundation-model checkpoint, the per-window scaling applied at downstream evaluation time matches the one used during its own pre-training.
For the $58$ checkpoints of the core grid, a per-window robust scaler divides every channel by the median, taken across channels, of the per-channel standard deviations; values exceeding $\pm 15$ standard deviations are finally clipped.
For the public REVE baseline, we apply per-window the standardisation described in the REVE paper~\citep{ouahidiREVEFoundationModel2025}: z-score followed by clipping at $\pm 15$ standard deviations.
The three non-foundation baselines (EEGNet, EEGConformer, ShallowFBCSPNet) are trained from scratch on each downstream task and receive the raw signals in microvolts, with no per-window normalisation.

\subsection{Downstream pipeline of foundation models}
\label{app:downstream-fm}

For every (checkpoint, dataset) pair, the downstream pipeline of a foundation model operates in three stages: frozen feature extraction, Gaussian random projection, and a closed-form ridge solve.
There is no stochastic gradient descent.

\paragraph{Frozen feature extractor.}
For every input window, the encoder is operated on full, unmasked inputs as a frozen feature extractor.
It outputs the post-context tokens $\mathbf{Z}\in\mathbb{R}^{B\times C\times T\times d}$ produced by its last encoder layer, where $B$ is the batch size, $C$ the number of channels of the downstream dataset, $T$ the number of temporal patches per window, and $d=512$ the latent dimension.
No trainable parameter is updated by the downstream protocol.

\paragraph{Per-dataset feature dimensionality before the random projection.}
\label{app:oeb-feature-dims}
The number of features $D_\mathrm{orig}$ that the ridge probe sees per example, before the Gaussian random projection described next, is determined entirely by the encoder's pre-head tensor shape: both the REVE baseline and our encoder emit a $(B, n_\mathrm{chans}, n_\mathrm{patches}, 512)$ tensor that is flattened to a single vector per example.
With the shared tokenisation parameters --- patch length of $200$ samples, $20$-sample overlap (so a stride of $180$ samples), and embedding dimension $512$ --- the number of temporal patches is $n_\mathrm{patches}=\lfloor (n_\mathrm{times}-200)/180\rfloor + 1$ and the feature count is
\begin{equation}
  D_\mathrm{orig} \;=\; n_\mathrm{chans}\cdot n_\mathrm{patches}\cdot 512.
  \label{eq:D-orig}
\end{equation}

Table~\ref{tab:oeb-feature-dims} reports $D_\mathrm{orig}$ on each of the $12$ OpenEEGBench datasets, computed from the per-dataset window shapes $(n_\mathrm{chans}, n_\mathrm{times})$.

\begin{table}[t]
  \centering
  \caption{Per-dataset feature dimension $D_\mathrm{orig}$ before the Gaussian random projection, computed from Eq.~\eqref{eq:D-orig}. Columns $n_\mathrm{chans}$ and $n_\mathrm{times}$ are read from the Hugging Face Zarr metadata of each dataset; $n_\mathrm{patches}$ follows from the shared tokenisation. The last column flags whether the projection step activates (always true here, since $D_\mathrm{orig}>5{,}000$ on every cell).}
  \label{tab:oeb-feature-dims}
  \footnotesize
  \begin{tabular}{lrrrrc}
    \toprule
    Dataset & $n_\mathrm{chans}$ & $n_\mathrm{times}$ & $n_\mathrm{patches}$ & $D_\mathrm{orig}$ & projected? \\
    \midrule
    \texttt{arithmetic\_zyma2019} & $19$ & $1{,}000$ & $5$  & $48{,}640$  & yes \\
    \texttt{bcic2a}               & $22$ & $800$    & $4$  & $45{,}056$  & yes \\
    \texttt{bcic2020-3}           & $64$ & $600$    & $3$  & $98{,}304$  & yes \\
    \texttt{physionet}            & $64$ & $600$    & $3$  & $98{,}304$  & yes \\
    \texttt{chbmit}               & $17$ & $2{,}000$ & $11$ & $95{,}744$  & yes \\
    \texttt{faced}                & $26$ & $2{,}000$ & $11$ & $146{,}432$ & yes \\
    \texttt{isruc-sleep}          & $6$  & $6{,}000$ & $33$ & $101{,}376$ & yes \\
    \texttt{mdd\_mumtaz2016}      & $19$ & $1{,}000$ & $5$  & $48{,}640$  & yes \\
    \texttt{seed-v}               & $62$ & $200$    & $1$  & $31{,}744$  & yes \\
    \texttt{seed-vig}             & $17$ & $1{,}600$ & $8$  & $69{,}632$  & yes \\
    \texttt{tuab}                 & $16$ & $2{,}000$ & $11$ & $90{,}112$  & yes \\
    \texttt{tuev}                 & $21$ & $1{,}000$ & $5$  & $53{,}760$  & yes \\
    \bottomrule
  \end{tabular}
\end{table}

\paragraph{Random projection and seeds.}
The flattened output tensor $\mathbf{Z}$ of dimension $D_\mathrm{orig}$ is then projected into a lower-dimensional space by Gaussian random projection $\mathbb{R}^{D_\mathrm{orig}}\!\to\!\mathbb{R}^{D_\mathrm{proj}}$ with $D_\mathrm{proj}=5{,}000$; its entries are drawn from $\mathcal{N}(0,\,1/D_\mathrm{proj})$.
This random projection is the only stochastic component of the foundation-model downstream pipeline (the closed-form ridge solve being deterministic given the projection); each seed therefore corresponds to an independent draw of the projection.
We use $5$ seeds for the final-epoch (epoch~$10$) checkpoint of every (framework, mask) configuration, so that the headline numbers reported in the main paper are summarised as a mean over $5$ seeds.
We use only $3$ seeds for the intermediate-epoch checkpoints used in the training-trajectory plots to save compute resources.
This choice is reasonable because only the final-epoch checkpoints ($5$ seeds) are used in statistical comparisons; the training-trajectory curves  ($3$ seeds) are meant to be used for qualitative comparison only.

\paragraph{Closed-form ridge solve.}
The ridge solver accumulates sufficient statistics ($\mathbf{Z}_\mathrm{proj}^{\top}\mathbf{Z}_\mathrm{proj}$, $\mathbf{Z}_\mathrm{proj}^{\top}Y$, mean and second-moment sums) in double precision over the training split.
The centred $D_\mathrm{proj}\times D_\mathrm{proj}$ correlation matrix is eigendecomposed once, and a fixed log-spaced grid of $17$ values $\lambda\in\{10^{-8}, 10^{-7}, \ldots, 10^{8}\}$ is swept in the eigenbasis to pick the regulariser that maximises the validation metric (balanced accuracy for classification datasets, $R^{2}$ for regression datasets); ties are broken in favour of the largest $\lambda$.

\subsection{Downstream pipeline of non-foundation models}
\label{app:downstream-non-fm}

The three non-foundation baselines (EEGNet~\citep{lawhernEEGNetCompactConvolutional2018}, EEGConformer~\citep{songEEGConformerConvolutional2023a}, and ShallowFBCSPNet~\citep{schirrmeisterDeepLearningConvolutional2017a}) are trained from scratch on each downstream dataset using the OpenEEGBench per-task training protocol.

\paragraph{End-to-end training.}
The three non-foundation baselines are optimised with AdamW at a peak learning rate of $10^{-3}$ and a batch size of $64$, under a cosine learning-rate schedule decaying from the peak rate down to $\eta_\mathrm{min}=10^{-6}$ over the full training duration.
Each model is trained for at most $30$ epochs with early stopping on the validation loss (patience $10$), and gradients are clipped to a norm of $1.0$.
The same hyperparameters are used across all $12$ downstream datasets and the three architectures, with no per-dataset overrides.

\paragraph{Seeds.}
Each seed independently re-initialises the backbone weights and the linear-head weights, re-shuffles the training batches, so the reported variances reflect both initialisation and training-time stochasticity.

\subsection{Cross-dataset score normalisation}\label{app:score-normalisation}

The 12 OpenEEGBench datasets are scored with different metrics
(balanced accuracy for the 11 classification tasks, $R^{2}$ for the
\texttt{seed-vig} regression task) and on different absolute scales,
so the raw scores cannot be averaged directly across datasets. To
report a single cross-dataset summary (the \emph{Average (norm.)} row
  of \autoref{fig:oeb-v9-per-dataset} and the $y$-axis of
\autoref{fig:oeb-timecourse-per-dataset}), we linearly map the raw score $s$
on dataset $d$ to a dataset-normalised score
\begin{equation}
  \tilde{s}^{d} \;=\; \frac{s - s^{d}_{\min}}{s^{d}_{\max} - s^{d}_{\min}},
  \label{eq:score-normalisation}
\end{equation}
where $s^{d}_{\min}$ and $s^{d}_{\max}$ are the minimum and maximum,
across the $58$ sweep configurations ($29$ mask cells $\times\,2$
frameworks), of the seed-averaged score on dataset $d$.
By construction, the worst sweep configuration of each dataset maps to
$\tilde{s}^{d}=0$ and the best to $\tilde{s}^{d}=1$, so the resulting
quantity is comparable across datasets and the cross-dataset average
weights every dataset uniformly. \autoref{tab:score-normalisation}
reports the constants $s^{d}_{\min}$ and $s^{d}_{\max}$ used throughout
the paper.

\begin{table}[t]
  \centering
  \caption{Per-dataset normalisation constants $s^{d}_{\min}$ and
    $s^{d}_{\max}$ used in \autoref{eq:score-normalisation}. They are
    the minimum and maximum, across the $58$ v9 sweep configurations,
    of the seed-averaged score on each dataset (balanced accuracy or
    $R^{2}$). The negative values for \texttt{seed-vig} are an instance
    of the negative-$R^{2}$ phenomenon discussed in
  \autoref{app:seed-vig-r2}.}
  \label{tab:score-normalisation}
  \footnotesize
  \begin{tabular}{lcc}
\toprule
 & min & max \\
Dataset &  &  \\
\midrule
arithmetic\_zyma2019 & 0.595 & 0.778 \\
bcic2020-3 & 0.231 & 0.294 \\
bcic2a & 0.263 & 0.475 \\
chbmit & 0.771 & 0.922 \\
faced & 0.158 & 0.332 \\
isruc-sleep & 0.497 & 0.720 \\
mdd\_mumtaz2016 & 0.698 & 0.864 \\
physionet & 0.326 & 0.585 \\
seed-v & 0.263 & 0.318 \\
seed-vig & -0.848 & -0.082 \\
tuab & 0.706 & 0.824 \\
tuev & 0.792 & 0.949 \\
\bottomrule
\end{tabular}

\end{table}

\subsection{SEED-VIG: negative \texorpdfstring{$R^{2}$}{R²} under ridge probing}\label{app:seed-vig-r2}

\texttt{seed-vig} is the only regression dataset in OpenEEGBench: the target
is the per-window PERCLOS vigilance score in $[0,1]$, and the test metric is
\texttt{sklearn.metrics.r2\_score}. Across all our checkpoints, ridge probing
on \texttt{seed-vig} returns a \emph{negative} $R^{2}$ around $-0.35$. This
is not a bug: it is a structural consequence of the predefined subject split
combined with the closed-form ridge solver, and the score still ranks models
meaningfully above a constant baseline.

\begin{figure}[t]
  \centering
  \includegraphics[scale=1]{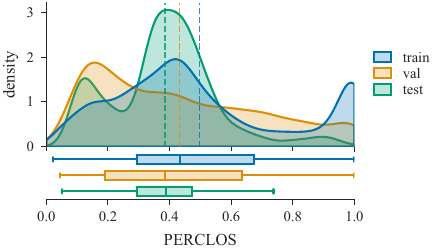}
  \caption{Per-window PERCLOS distribution on the OpenEEGBench
    \texttt{seed-vig} split (subject-based: train${=}\{1,2,3,10,\ldots,21\}$,
    val${=}\{4,5\}$, test${=}\{6,7,8,9\}$). \emph{Top:} Kernel-density
    estimate per split; the dashed verticals mark the per-split means.
    \emph{Bottom:} matching horizontal box plots (medians, IQR, whiskers
    at $1.5\times$IQR; outliers hidden). The three splits differ both in
    location (means drift monotonically from train to test) and in shape:
    train and val are broad and have a heavy right tail near $\mathrm{PERCLOS}\!\approx\!1$,
    whereas test is unimodal, sharply concentrated around the central
    range, and lacks the high-PERCLOS tail entirely. This is what the
  closed-form ridge probe cannot recover from.}
  \label{fig:seed-vig-label-distribution}
\end{figure}

The distribution shift across the OpenEEGBench split is shown in Figure~\ref{fig:seed-vig-label-distribution}
and quantified in Table~\ref{tab:seed-vig-shift}.The test distribution is much narrower than
either train or val and is missing the heavy right tail near $1$ that
both training-side splits carry.

\begin{table}[t]
  \centering
  \caption{Per-window PERCLOS statistics for the OpenEEGBench
    \texttt{seed-vig} split. The four test subjects happen to be both less
    drowsy on average and more uniform than the training subjects, which
  creates the negative-$R^{2}$ floor described in the text.}
  \label{tab:seed-vig-shift}
  \footnotesize
  \begin{tabular}{lrrrr}
    \toprule
    Split & $n_\mathrm{windows}$ & mean & std & range \\
    \midrule
    train & $13{,}275$ & $0.498$ & $0.283$ & $[0.022,\,1.000]$ \\
    val   & $3{,}540$  & $0.433$ & $0.269$ & $[0.044,\,1.000]$ \\
    test  & $3{,}540$  & $0.385$ & $0.175$ & $[0.050,\,1.000]$ \\
    \bottomrule
  \end{tabular}
\end{table}

Both pathologies hurt the ridge probe in the same direction. The
closed-form solver fixes its bias to
$b = \bar{y}_\mathrm{train} - W^{\top} \bar{h}_\mathrm{train}$, so for any
feature mapping with limited predictive power on the new subjects the
predictions concentrate near $\bar{y}_\mathrm{train}=0.498$, i.e.\ around
the right-most dashed line in
Figure~\ref{fig:seed-vig-label-distribution}, while the test distribution
is centred on the left-most one. The mean mismatch alone contributes
\[
  \frac{(\bar{y}_\mathrm{train}-\bar{y}_\mathrm{test})^{2}}
  {\mathrm{Var}(y_\mathrm{test})}
  \;=\; \frac{0.113^{2}}{0.175^{2}} \;\approx\; 0.413
\]
to $1-R^{2}$, setting a hard floor of $R^{2}\!\approx\!-0.413$ for any
near-constant predictor. Selecting $\lambda$ on val cannot rescue this:
the val box in Figure~\ref{fig:seed-vig-label-distribution} has roughly
the same width as train and only a small mean offset, so the val-side
penalty is mild and the probe receives no signal that the test box has
collapsed to roughly a third of its width.

To make this floor concrete we report the score of two reference
predictors in Table~\ref{tab:seed-vig-r2-floor}: a
\texttt{DummyRegressor(strategy=\textquotesingle mean\textquotesingle)} that always returns
$\bar{y}_\mathrm{train}$, and the REVE ridge-probe baseline (frozen,
ridge probe with $5$ projection seeds).

\begin{table}[t]
  \centering
  \caption{Test $R^{2}$ on \texttt{seed-vig} for the train-mean dummy
    predictor and for the frozen REVE baseline under the OpenEEGBench
    ridge probe (5 random-projection seeds). The dummy score is the
    structural floor implied by the subject split visible in
    Figure~\ref{fig:seed-vig-label-distribution}; the ridge probe sits
    above it, indicating that the frozen features carry information
  about PERCLOS even though the metric stays negative.}
  \label{tab:seed-vig-r2-floor}
  \footnotesize
  \begin{tabular}{lr}
    \toprule
    Predictor & test $R^{2}$ \\
    \midrule
    \texttt{DummyRegressor} (predicts $\bar{y}_\mathrm{train}=0.498$) & $-0.413$ \\
    REVE baseline, ridge probe (mean over 5 seeds) & $-0.355\,\pm\,0.016$ \\
    \midrule
    Gap above the dummy floor & $+0.058$ \\
    \bottomrule
  \end{tabular}
\end{table}

The frozen REVE features therefore lift the score by $\approx\!0.06$
$R^{2}$ above the constant-prediction floor, which is the relevant
quantity when comparing checkpoints on this dataset. We report the raw
$R^{2}$ throughout the paper for consistency with the rest of OpenEEGBench,
but the reader should keep in mind that on \texttt{seed-vig} the
informative axis is the gap to the $-0.413$ floor rather than the sign
of the score itself.

\section{Spatial vs.\ temporal redundancy of the EEG signal}\label{app:signal-correlation}

\begin{figure}[t]
  \centering
  \includegraphics[scale=1]{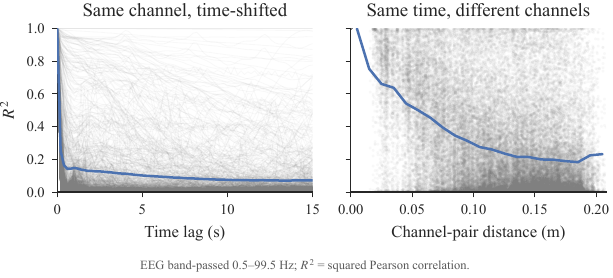}
  \caption{Squared Pearson correlation $R^{2}$ of the broadband EEG on a random subsample of REVE windows.
    \emph{Left:} within-channel autocorrelation as a function of the time-lag (grey lines: per-window curves; coloured line: average across windows).
    \emph{Right:} between-channel correlation as a function of the Euclidean inter-electrode distance (grey points: per-pair $R^{2}$; coloured line: distance-binned average).
  Temporal redundancy decays much faster than spatial redundancy.}
  \label{fig:signal-correlation}
\end{figure}

The two axes along which we mask, channels and time, are not interchangeable in the underlying signal.
\autoref{fig:signal-correlation} reports the squared Pearson correlation $R^{2}$ of the broadband EEG between (\emph{left}) the same channel at two different time points and (\emph{right}) two different channels at the same time point, measured on a random subsample of REVE windows.
Within-channel autocorrelation drops sharply with the time-lag, falling below $R^{2}\approx 0.2$ within $\sim\!2\,$s; between-channel correlation, in contrast, decays much more slowly with electrode distance and remains non-negligible across the entire scalp.
A $1\,$s temporal patch of a given channel can therefore be predicted reasonably well from its own neighbouring patches alone, whereas channel-wise spatial structure is genuinely shared between channels and is harder to recover from local temporal context.
This anchors the qualitative interpretation of the masking grid used in Section~\ref{sec:results}.

\section{Bias-inflation collapse}\label{app:bias-inflation-collapse}

This appendix gives the full mechanistic analysis of \emph{bias-inflation collapse}, the JEPA-only failure mode at $r=\texttt{"all"}$ described in \nameref{par:finding3} and which, to our knowledge, has not been documented in the prior literature.
We characterise the phenomenon geometrically (\autoref{app:bic-notation}), document its empirical signature (\autoref{app:bic-empirical}), explain why it is the optimum of the JEPA loss at $r=\texttt{"all"}$ but not at smaller radii nor under MAE (\autoref{app:bic-mechanism}), show why none of the standard collapse detectors fire on it (\autoref{app:bic-vicreg}), discuss why an in-training detector for this regime is non-trivial in the multi-montage pre-training setting (\autoref{app:bic-detection-challenges}), list mechanistic hypotheses we ruled out during the investigation (\autoref{app:bic-failed}), and situate the phenomenon within the broader literature on self-supervised collapse (\autoref{app:bic-related-work}).

\subsection{Phenomenon and notation}\label{app:bic-notation}

Let the encoder produce, at the end of a forward pass, a feature tensor $z(n, c, t) \in \mathbb{R}^{d}$ for every window $n$, non-padded channel $c$, and time-patch $t$.
We decompose every token into its window-mean component and its input-driven residual:
\begin{equation}\label{eq:bic-decomposition}
  z(n, c, t) \;=\; \mu(c, t) \;+\; \delta(n, c, t),
  \qquad
  \mu(c, t) \;:=\; \mathbb{E}_n\,z(n, c, t).
\end{equation}
We track two scalars summarising this decomposition: the bias magnitude $\|\mu\|$, defined as the average of $\|\mu(c, t)\|$ over $(c, t)$, and the input-driven magnitude $\sigma_\text{token} := \mathbb{E}_{n,c,t}\,\|\delta(n, c, t)\|$.
The main empirical claim is that, across pre-training of JEPA at $r=\texttt{"all"}$, $\|\mu\|$ inflates by a factor of $2$--$3$ while $\sigma_\text{token}$ collapses by a factor of $5$--$10$, yielding tokens dominated by a fixed per-channel-position bias and carrying vanishing input-driven information.
Healthy ($r \le 12\,\text{cm}$) JEPA runs and the MAE control at $r=\texttt{"all"}$ both keep $\|\mu\|$ approximately stable and $\sigma_\text{token}$ growing.

\subsection{Empirical signature}\label{app:bic-empirical}

We measure the components of \autoref{eq:bic-decomposition} on the data received by the probe at downstream evaluation (a single subject of physionet and bcic2a, $84$ and $576$ windows respectively, processed through $10$ representative pre-trained checkpoints --- the $5$ JEPA $r$-sweep cells at $L=2$, the $4$ JEPA $L$-sweep cells at $r=\texttt{"all"}$, and the MAE $r=\texttt{"all"}$ control) at every pre-training epoch (epochs $1, \ldots, 10$).
Four scalars summarise each (run, epoch) cell: $\|\mu\|$, $\sigma_\text{token}$, the dimensionless ratio $\sigma_\text{token}/\sigma_\text{chan}$ where $\sigma_\text{chan}:=\mathbb{E}_{n,t}\,\|z(n,c,t) - \mathbb{E}_c z(n,c,t)\|$ measures the spread of $z$ across channels at fixed $(n, t)$, and the mean intra-channel cosine similarity $\overline{\cos_\text{intra}}$ averaged over $(c, t)$ of unit-normalised tokens across windows.

\begin{figure}[t]
  \centering
  \includegraphics[width=\linewidth]{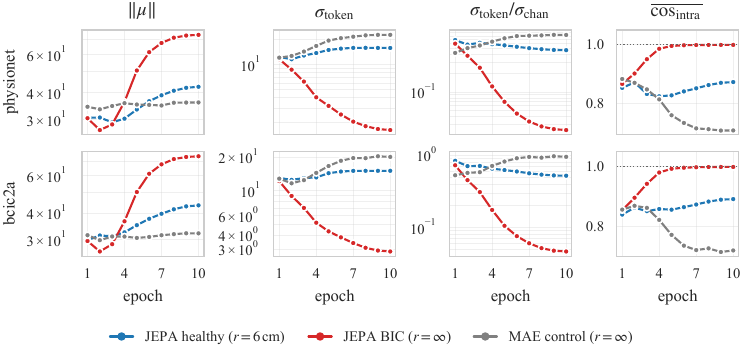}
  \caption{
    Bias-inflation collapse signature on OEB probe data, cross-window metrics computed on a single subject of physionet (top) and bcic2a (bottom) processed through each frozen checkpoint at every pre-training epoch.
    Only JEPA at $r=\texttt{"all"}$ shows the simultaneous bias inflation ($\|\mu\|\uparrow$), residual collapse ($\sigma_\text{token}\downarrow$), ratio drop and direction collapse ($\overline{\cos_\text{intra}}\!\to\!1$) predicted by the lookup-table mechanism (\autoref{app:bic-mechanism}).
    Healthy JEPA ($r=6\,$cm) and the MAE control ($r=\texttt{"all"}$) remain stable on all four metrics.
  }
  \label{fig:bic-trajectories}
\end{figure}

The four trajectories tell a coherent story: only JEPA at $r=\texttt{"all"}$ exhibits the simultaneous bias inflation, residual collapse, ratio collapse, and asymptote of the cosine similarity at $1.0$. Every window of a given channel produces essentially the same direction in feature space.
JEPA at $r=6\,$cm and MAE at $r=\texttt{"all"}$ remain stable on all four metrics across pre-training.

\paragraph{Phase transition in $r$.} The dependence on the spatial radius is sharp rather than smooth.
Sweeping the JEPA grid at $L=2$ and the final epoch across $r \in \{\texttt{"one"}, 6, 9, 12\,\text{cm}, \texttt{"all"}\}$, the ratio $\sigma_\text{token}/\sigma_\text{chan}$ stays in $[0.36, 0.56]$ across the four finite-radius cells and then drops to $\approx 0.024$ on physionet and $\approx 0.047$ on bcic2a at $r=\texttt{"all"}$ --- a phase transition of one order of magnitude in both datasets.
The dependence on the mask ratio $\rho$, which the core grid holds fixed, is examined in \autoref{app:ablation-rho}: $\rho$ influences the collapsed cell more than any other, and the cell improves at higher $\rho$ without becoming competitive.

\begin{figure}[t]
  \centering
  \includegraphics[scale=1]{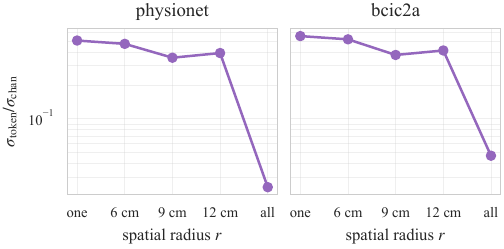}
  \caption{
    Phase transition in the spatial radius $r$ at $L=2$ and the final epoch.
    The ratio $\sigma_\text{token}/\sigma_\text{chan}$ stays clustered around $0.4$ across the four finite-radius cells and drops by an order of magnitude at $r=\texttt{"all"}$ on both physionet and bcic2a --- a sharp transition, not a smooth trend.
  }
  \label{fig:bic-r-sweep}
\end{figure}

\paragraph{Predictivity across the full grid.} Aggregating across $10$ configurations (full r-sweep at $L=2$, full $L$-sweep at $r=\texttt{"all"}$, and the MAE $r=\texttt{"all"}$ control) on both physionet and bcic2a, the log of $\sigma_\text{token}/\sigma_\text{chan}$ at the final epoch predicts the final downstream score with Pearson $r = +0.95$ on physionet and $+0.91$ on bcic2a.
The bias-inflation signature is therefore not a quirk of one configuration but a quantitative predictor of downstream collapse across the full $(L, r)$ grid.

\begin{figure}[t]
  \centering
  \includegraphics[scale=1]{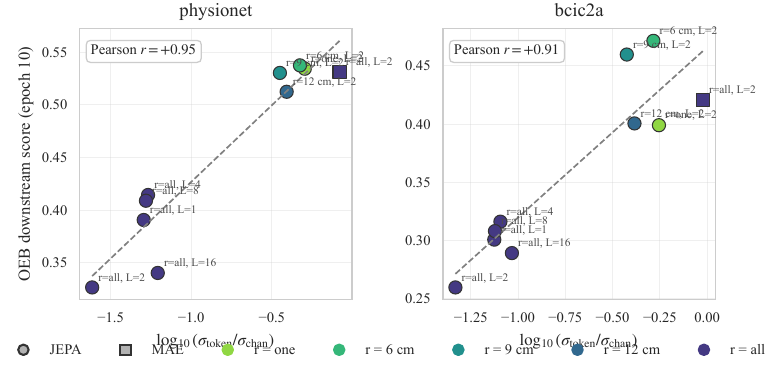}
  \caption{
    Predictivity of $\log_{10}(\sigma_\text{token}/\sigma_\text{chan})$ for the downstream OEB score at the final epoch across the $10$ (framework, $r$, $L$) configurations.
    Each point is one frozen checkpoint at epoch $10$.
    Markers: framework ($\circ$ JEPA, $\square$ MAE).
    Colours: spatial radius (light $\to$ dark for small $\to$ large $r$).
    Pearson $r=+0.95$ on physionet and $+0.91$ on bcic2a.
  }
  \label{fig:bic-scatter}
\end{figure}

\paragraph{Multi-subject robustness.} We replicate the four metrics on five different subjects of each downstream dataset.
Healthy and MAE trajectories are essentially subject-invariant; the bias-inflation trajectory is also robust, with subject-to-subject variation contained within $5\%$ of the run-level mean.

\subsection{Spectral selectivity of the collapsed representation}\label{app:bic-spectral}

To characterise what the collapsed representation retains, we fit ridge probes on the frozen features of all $58$ configurations of the core grid, across all twelve OpenEEGBench datasets, to predict the per-channel log band-power in the $\delta$, $\theta$, $\alpha$, $\beta$ and $\gamma$ bands.
\autoref{tab:bic-bandpower} reports the decodability of each band (mean over channels), per dataset, for the two optima of the grid, the collapsed cell and an untrained encoder used as a floor.
Bias-inflation collapse is spectrally selective, and this replicates on $12$ of $12$ datasets: the collapsed JEPA cell ($r=\texttt{"all"}$, $L=2$) retains the high bands far better than $\delta$, $\theta$ and $\alpha$, and its low-band decodability is the lowest of the grid on most datasets.
The collapsed model therefore does not decode nothing: it keeps broadband amplitude and loses the slow structure of the signal.

\begin{table}[!h]
  \centering
  \caption{Band-power decodability from frozen final-epoch features ($R^2$, mean over channels), per frequency band and per OpenEEGBench dataset, for the two optima of the core grid, the collapsed JEPA cell and an untrained encoder.}
  \label{tab:bic-bandpower}
  \scriptsize
  \renewcommand{\arraystretch}{0.88}
  \begin{tabular}{lrrrr}
    \toprule
    Band / dataset & MAE $(9\,\text{cm}, L{=}2)$ & JEPA $(9\,\text{cm}, L{=}2)$ & JEPA $(\texttt{"all"}, L{=}2)$, collapsed & Untrained \\
    \midrule
    \multicolumn{5}{l}{\textbf{$\delta$ band}} \\
    \quad \texttt{arithmetic\_zyma2019} & $0.72$ & $0.50$ & $-0.02$ & $0.49$ \\
    \quad \texttt{bcic2020-3} & $0.62$ & $0.40$ & $0.11$ & $0.30$ \\
    \quad \texttt{bcic2a} & $0.70$ & $0.47$ & $-0.08$ & $0.29$ \\
    \quad \texttt{chbmit} & $0.84$ & $0.75$ & $0.04$ & $0.67$ \\
    \quad \texttt{faced} & $0.67$ & $0.51$ & $0.03$ & $0.42$ \\
    \quad \texttt{isruc-sleep} & $0.82$ & $0.64$ & $-0.05$ & $0.60$ \\
    \quad \texttt{mdd\_mumtaz2016} & $0.75$ & $0.63$ & $0.04$ & $0.52$ \\
    \quad \texttt{physionet} & $0.75$ & $0.57$ & $0.02$ & $0.50$ \\
    \quad \texttt{seed-v} & $0.68$ & $0.49$ & $0.13$ & $0.33$ \\
    \quad \texttt{seed-vig} & $0.66$ & $0.50$ & $0.03$ & $0.46$ \\
    \quad \texttt{tuab} & $0.77$ & $0.63$ & $0.05$ & $0.48$ \\
    \quad \texttt{tuev} & $0.80$ & $0.68$ & $0.14$ & $0.57$ \\
    \midrule
    \multicolumn{5}{l}{\textbf{$\theta$ band}} \\
    \quad \texttt{arithmetic\_zyma2019} & $0.54$ & $0.44$ & $-0.06$ & $0.17$ \\
    \quad \texttt{bcic2020-3} & $0.63$ & $0.55$ & $0.24$ & $0.20$ \\
    \quad \texttt{bcic2a} & $0.59$ & $0.45$ & $0.10$ & $0.08$ \\
    \quad \texttt{chbmit} & $0.73$ & $0.68$ & $-0.01$ & $0.40$ \\
    \quad \texttt{faced} & $0.60$ & $0.53$ & $0.09$ & $0.25$ \\
    \quad \texttt{isruc-sleep} & $0.69$ & $0.65$ & $-0.12$ & $0.19$ \\
    \quad \texttt{mdd\_mumtaz2016} & $0.77$ & $0.70$ & $-0.01$ & $0.42$ \\
    \quad \texttt{physionet} & $0.66$ & $0.59$ & $0.05$ & $0.25$ \\
    \quad \texttt{seed-v} & $0.54$ & $0.48$ & $0.11$ & $0.06$ \\
    \quad \texttt{seed-vig} & $0.64$ & $0.59$ & $0.13$ & $0.35$ \\
    \quad \texttt{tuab} & $0.73$ & $0.68$ & $0.07$ & $0.37$ \\
    \quad \texttt{tuev} & $0.75$ & $0.71$ & $0.10$ & $0.39$ \\
    \midrule
    \multicolumn{5}{l}{\textbf{$\alpha$ band}} \\
    \quad \texttt{arithmetic\_zyma2019} & $0.71$ & $0.74$ & $0.17$ & $0.40$ \\
    \quad \texttt{bcic2020-3} & $0.68$ & $0.64$ & $0.24$ & $0.24$ \\
    \quad \texttt{bcic2a} & $0.73$ & $0.73$ & $0.07$ & $0.16$ \\
    \quad \texttt{chbmit} & $0.61$ & $0.65$ & $0.08$ & $0.25$ \\
    \quad \texttt{faced} & $0.58$ & $0.60$ & $0.07$ & $0.15$ \\
    \quad \texttt{isruc-sleep} & $0.72$ & $0.75$ & $-0.14$ & $0.34$ \\
    \quad \texttt{mdd\_mumtaz2016} & $0.67$ & $0.71$ & $-0.04$ & $0.34$ \\
    \quad \texttt{physionet} & $0.65$ & $0.66$ & $0.08$ & $0.20$ \\
    \quad \texttt{seed-v} & $0.52$ & $0.56$ & $0.03$ & $0.05$ \\
    \quad \texttt{seed-vig} & $0.69$ & $0.67$ & $0.12$ & $0.33$ \\
    \quad \texttt{tuab} & $0.74$ & $0.73$ & $0.12$ & $0.41$ \\
    \quad \texttt{tuev} & $0.68$ & $0.70$ & $0.15$ & $0.28$ \\
    \midrule
    \multicolumn{5}{l}{\textbf{$\beta$ band}} \\
    \quad \texttt{arithmetic\_zyma2019} & $0.30$ & $0.43$ & $0.23$ & $0.02$ \\
    \quad \texttt{bcic2020-3} & $0.62$ & $0.56$ & $0.36$ & $0.22$ \\
    \quad \texttt{bcic2a} & $0.36$ & $0.45$ & $0.32$ & $0.05$ \\
    \quad \texttt{chbmit} & $0.71$ & $0.76$ & $0.49$ & $0.41$ \\
    \quad \texttt{faced} & $0.61$ & $0.63$ & $0.38$ & $0.31$ \\
    \quad \texttt{isruc-sleep} & $0.66$ & $0.69$ & $-0.19$ & $0.36$ \\
    \quad \texttt{mdd\_mumtaz2016} & $0.66$ & $0.73$ & $0.54$ & $0.32$ \\
    \quad \texttt{physionet} & $0.49$ & $0.56$ & $0.34$ & $0.10$ \\
    \quad \texttt{seed-v} & $0.55$ & $0.60$ & $0.37$ & $0.16$ \\
    \quad \texttt{seed-vig} & $0.57$ & $0.54$ & $0.26$ & $0.28$ \\
    \quad \texttt{tuab} & $0.70$ & $0.69$ & $0.26$ & $0.40$ \\
    \quad \texttt{tuev} & $0.63$ & $0.68$ & $0.36$ & $0.31$ \\
    \midrule
    \multicolumn{5}{l}{\textbf{$\gamma$ band}} \\
    \quad \texttt{arithmetic\_zyma2019} & $0.15$ & $0.28$ & $0.32$ & $0.02$ \\
    \quad \texttt{bcic2020-3} & $0.51$ & $0.47$ & $0.42$ & $0.16$ \\
    \quad \texttt{bcic2a} & $0.51$ & $0.67$ & $0.69$ & $0.07$ \\
    \quad \texttt{chbmit} & $0.68$ & $0.74$ & $0.62$ & $0.46$ \\
    \quad \texttt{faced} & $0.59$ & $0.68$ & $0.60$ & $0.35$ \\
    \quad \texttt{isruc-sleep} & $0.59$ & $0.64$ & $0.20$ & $0.29$ \\
    \quad \texttt{mdd\_mumtaz2016} & $0.42$ & $0.52$ & $0.24$ & $0.15$ \\
    \quad \texttt{physionet} & $0.36$ & $0.50$ & $0.53$ & $0.05$ \\
    \quad \texttt{seed-v} & $0.56$ & $0.66$ & $0.62$ & $0.13$ \\
    \quad \texttt{seed-vig} & $0.51$ & $0.56$ & $0.36$ & $0.23$ \\
    \quad \texttt{tuab} & $0.64$ & $0.66$ & $0.37$ & $0.37$ \\
    \quad \texttt{tuev} & $0.55$ & $0.63$ & $0.49$ & $0.27$ \\
    \bottomrule
  \end{tabular}
\end{table}

\subsection{The lookup-table mechanism}\label{app:bic-mechanism}

The structure of the JEPA loss admits, at $r=\texttt{"all"}$, a particularly simple trivial fixed point.
Recall that the JEPA loss
\[
  \mathcal{L}_\text{JEPA}
  \;=\;
  \frac{1}{|\mathcal{M}|\,d}
  \sum_{(c,t)\in\mathcal{M}}
  \bigl\|\hat{z}_{c,t} - \bar{z}_{c,t}\bigr\|_{2}^{2}
\]
is minimised whenever the predictor's output $\hat{z}_{c,t}$ matches the EMA teacher's output $\bar{z}_{c,t}$ at every masked position.
Any function that depends only on the position $(c, t)$ is a valid joint solution, since both student and teacher pass through the same positional encoder and the same architecture (modulo the EMA delay).
We call such a solution a \emph{lookup table}: a deterministic mapping $(c, t) \mapsto M(c, t) \in \mathbb{R}^{d}$ that the encoder learns to emit identically across windows, regardless of the input EEG.
If the encoder converges to this solution then $z(n, c, t) \approx M(c, t)$ for every $n$, the EMA teacher does likewise, and the predictor (a small transformer that receives the positional embeddings of the masked tokens together with the encoder's contextual output) can reproduce $M(c, t)$ from positions alone.
The JEPA loss reaches a near-zero floor without the encoder ever encoding any input-driven information.

\paragraph{Why $r=\texttt{"all"}$ unlocks the shortcut.} The competing non-trivial solution is for the encoder to exploit the redundancy of EEG signals across spatial neighbours: when channel $c$ at time-patch $t$ is masked but a nearby channel $c'$ at the same $t$ is not, volume-conducted activity at $c'$ is strongly informative of the content at $c$.
At $r \le 12\,\text{cm}$ this same-time spatial neighbour is preserved in the predictor's context for nearly every masked position, and the spatial-redundancy path is at least as attractive as the lookup table.
At $r=\texttt{"all"}$, every channel of the masked time-patch is masked together, so no same-time spatial neighbour ever survives in the context.
The only signal left is the temporal axis, which is poorly predictive of EEG content at the timescale of our window (\nameref{par:spat-temp-redundancy}).
The lookup table becomes the strictly easier path, and the optimiser converges to it.

\paragraph{Why MAE forbids the shortcut structurally.} Under MAE, the target at masked position $(c, t)$ is the raw input patch $x_{c, t} \in \mathbb{R}^{p}$, and the readout is a fixed linear map $W \in \mathbb{R}^{p \times d}$ followed by an $L_2$ loss.
If the encoder converged to the same lookup-table solution $z(n, c, t) \approx M(c, t)$, the readout $W \cdot M(c, t)$ would be a per-position constant, identical across windows; the loss would then equal the cross-window variance of the raw EEG patch, which is of order one and far from zero.
The reconstruction head therefore forbids the lookup-table solution and forces the encoder to keep multiple input-dependent latent directions alive --- even at $r=\texttt{"all"}$, where MAE merely under-performs but does not collapse.

\subsection{Why standard detectors fail to fire}\label{app:bic-vicreg}

Three classes of collapse detector documented in the JEPA literature fail to register bias-inflation collapse, including the three triggers of our collapse-detection callback (Appendix~\ref{app:jepa-details}).

\begin{itemize}
  \item \textbf{Per-dimension variance check.} The variance of $z_d$ across the batch remains well above the $\epsilon = 10^{-2}$ threshold throughout pre-training, because the inflated bias $\mu(c, t)$ varies across $(c, t)$ cells and contributes $\Theta(\|\mu\|^2)$ to the per-dimension variance.
  \item \textbf{Loss divergence.} The main JEPA loss does not blow up: in the lookup-table regime, student and teacher converge \emph{together} to the same $M(c, t)$, and the loss drives smoothly toward zero rather than past $10^{4}$.
  \item \textbf{Trivial-constant collapse.} The loss does not collapse to $< 10^{-3}$ within the first $500$ steps either, because the lookup table is a non-constant function of position and is learnt over many epochs (the bias inflation occurs gradually between epochs $4$ and $10$, see \autoref{app:bic-empirical}).
\end{itemize}

\paragraph{VICReg-style regularisers are unlikely to rescue the regime.} The VICReg variance hinge $\mathcal{L}_\text{var} = \frac{1}{d} \sum_{i=1}^{d} \max(0, \gamma - \sqrt{\operatorname{Var}(z_i)})$ is satisfied because $\operatorname{Var}(z_i)$ is dominated by the bias term and remains above any reasonable target $\gamma$.
The covariance term $\mathcal{L}_\text{cov} = \frac{1}{d}\sum_{i \ne j} \operatorname{Cov}(z_i, z_j)^2$ is satisfied whenever the lookup table $\{M(c, t)\}$ spreads isotropically across the $d$ feature dimensions, and we observe empirically that it does: the effective rank of $\{M(c, t)\}$ grows from $\sim 5$ at epoch $1$ to $\sim 25$ at epoch $10$ on physionet, indicating that the encoder actively distributes the lookup directions across feature dimensions.
The combined objective $\mathcal{L}_\text{JEPA} + \mathcal{L}_\text{var} + \mathcal{L}_\text{cov}$ therefore appears to be satisfied by the lookup table at no cost, so VICReg \citep{bardesVICRegVarianceInvarianceCovarianceRegularization2021}, C-JEPA \citep{moConnectingJointEmbeddingPredictive2024}, and similar variance-preserving regularisers are unlikely to detect or prevent bias-inflation collapse in the multichannel setting.
We do not test this prediction empirically: our entire grid is run in the no-auxiliary-loss configuration to keep the framework axis at strict parity with MAE, so adding VICReg or C-JEPA at $r=\texttt{"all"}$ is left to future work.
Distributional regularisers that constrain the full embedding distribution rather than its second moments --- such as the SIGReg term of LeJEPA \citep{balestrieroLeJEPAProvableScalable2025} --- might in principle rule out the lookup-table fixed point, but we have not verified this either.

\subsection{Why detection during pre-training is non-trivial}\label{app:bic-detection-challenges}

The cross-window signature of \autoref{app:bic-empirical} ($\sigma_\text{token}/\sigma_\text{chan}$, $\overline{\cos_\text{intra}}$) is a quantitative predictor of bias-inflation collapse, but it relies on a fixed channel set across windows.
This assumption holds on a downstream evaluation dataset but breaks in pre-training: REVE windows come from recordings with different channel layouts, so there is no canonical correspondence of channels across windows in a batch.
A practical in-training detector would therefore have to operate \emph{within} a single window (or a single recording) without channel-set assumptions, and would have to discriminate the early-pre-training plateau of a collapsed run from the legitimately slow take-off of a healthy run.
We could not construct a detector that meets both requirements reliably; this paragraph documents the natural candidates we explored and why each fell short, so that future work can build on the attempt.

\paragraph{Within-window candidate quantities.} Given the encoder output $z \in \mathbb{R}^{W \times C \times T \times d}$ on a batch of $W$ windows (with per-window non-padded channels indicated by a binary mask), the natural within-window analogues of the cross-window signature are
\begin{align*}
  \sigma_\text{time}
  &:= \operatorname*{mean}_{w}\;
  \operatorname*{mean}_{c\,\text{valid in } w}\;\;
  \operatorname*{mean}_{t}\;
  \bigl\|z_{w}(c, t) - \operatorname*{mean}_{t'}\,z_{w}(c, t')\bigr\|, \\
  \sigma_\text{chan}
  &:= \operatorname*{mean}_{w}\;
  \operatorname*{mean}_{t}\;\;
  \operatorname*{mean}_{c\,\text{valid in } w}\;
  \bigl\|z_{w}(c, t) - \operatorname*{mean}_{c'\,\text{valid in } w}\,z_{w}(c', t)\bigr\|,
\end{align*}
together with the effective rank $\operatorname{eff\_rank} := \exp\bigl(-\sum_i p_i \log p_i\bigr)$ of the centred token matrix flattened to $(N_\text{valid}\,T, d)$, computed per window and averaged across the batch.

\paragraph{What we observed on real REVE windows.} On a single $1$\,GB shard of the REVE training cache ($32$ windows per evaluation), we measured $\sigma_\text{time}/\sigma_\text{chan}$ and $\operatorname{eff\_rank}$ at four epochs ($1, 4, 7, 10$).
At epoch $1$ all three runs (healthy at $r=6\,$cm, BIC JEPA at $r=\texttt{"all"}$, MAE control at $r=\texttt{"all"}$) sit below the eventual healthy plateau on both metrics --- no static threshold would discriminate the BIC run from a healthy run at initialisation.
By epoch $4$ ($\sim\!30\%$ of the budget) the healthy and MAE runs have already taken off ($\sigma_\text{time}/\sigma_\text{chan} \approx 1.19$ and $0.96$ respectively, $\operatorname{eff\_rank}$ around $30$ and $36$), whereas the BIC run is still at its epoch-$1$ values ($\sigma_\text{time}/\sigma_\text{chan} \approx 0.49$, $\operatorname{eff\_rank} \approx 26$).

\begin{figure}[t]
  \centering
  \includegraphics[scale=1]{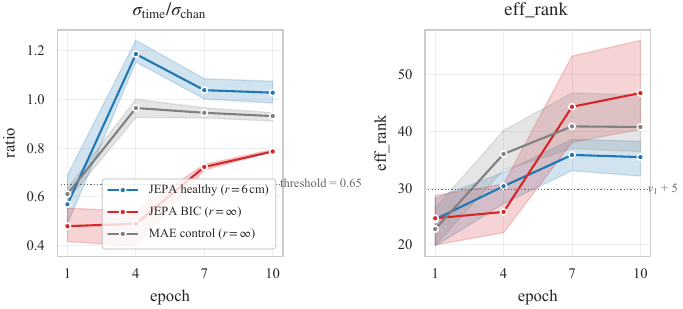}
  \caption{
    Candidate within-window proxies $\sigma_\text{time}/\sigma_\text{chan}$ (left) and $\operatorname{eff\_rank}$ (right) on real REVE pre-training windows, at epochs $1, 4, 7, 10$ for the three core runs.
    Lines: mean across the $32$ windows of the evaluation shard; bands: $25$--$75$ percentile.
    By epoch $4$ the BIC run is still at its epoch-$1$ values on both metrics, while healthy and MAE have already taken off --- but neither metric discriminates the runs at initialisation, and both can plausibly look low for legitimate reasons in early training.
  }
  \label{fig:bic-detector-reve}
\end{figure}

\paragraph{Why we do not turn this into a detector.} Although a conjunctive rule of the form ``$\sigma_\text{time}/\sigma_\text{chan} < 0.65$ \textsc{and} $\operatorname{eff\_rank} - \operatorname{eff\_rank}^{(1)} < 5$ after a warm-up of $\sim\!30\%$ of the budget'' separates the three runs in our shard test, we caution against using it as a stopping criterion in practice:
(i) its thresholds are calibrated on the same three runs that motivate the rule, with no independent validation set;
(ii) the warm-up window is itself a hyperparameter that depends on the planned epoch budget;
(iii) the underlying proxies are confounded with the legitimate slow take-off of healthy runs in the early epochs;
(iv) we have not tested the rule across other pre-training corpora, montage distributions, or backbones --- all of which could shift the healthy baseline.
A reliable in-training detector for bias-inflation collapse on multi-montage corpora most likely requires either a small held-out probe dataset evaluated periodically, or a distributional regulariser (such as the SIGReg term of LeJEPA \citep{balestrieroLeJEPAProvableScalable2025}) that structurally rules out the lookup-table fixed point; we leave both directions to future work.

\paragraph{Practical mitigation.} Until a reliable in-training detector exists, the simplest and most effective mitigation is to \emph{avoid the masking regime that triggers the collapse}.
Across the $58$-cell grid the failure is confined to JEPA at $r=\texttt{"all"}$ (\autoref{fig:oeb-grid}\,(C)), so the actionable rule for practitioners pre-training EEG foundation models with a JEPA-style pretext is to never mask every channel of the same time-patch jointly, and to prefer a moderate spatial radius ($r \in \{6, 9, 12\}\,\text{cm}$) as recommended in \nameref{par:finding2}.
Under this rule, no run in our grid exhibits bias-inflation collapse, and the question of in-training detection is sidestepped.

\subsection{Hypotheses we ruled out}\label{app:bic-failed}

For completeness, the following plausible mechanistic stories were ruled out by the data:

\begin{itemize}
  \item \emph{Rank-$1$ magnitude modulation,} $z = \alpha(n, c, t) \cdot u(c, t) + \varepsilon$: ruled out by measuring the parallel fraction of $\delta$ along $\mu$, which stays below $2\%$ across all runs --- $\delta$ is essentially orthogonal to $\mu$ rather than modulating its magnitude.
  \item \emph{Amplification of the random-init direction:} ruled out by $\cos\bigl(\mu(c, t)^{(10)}, \mu(c, t)^{(1)}\bigr) \approx 0.28$ for both healthy and collapsed runs --- the direction rotates as much during bias-inflation collapse as during healthy training, only the magnitude inflates differently.
  \item \emph{Topographic readout:} ruled out by the cosine-vs-distance Pearson correlation across channel pairs, which is \emph{lower} for collapsed runs ($\sim 0.40$) than for healthy ($\sim 0.55$) or MAE ($\sim 0.62$) --- the lookup table is essentially arbitrary across channels rather than smoothly topographic.
  \item \emph{Bias-rescue by feature standardisation:} ruled out by a probe-with-scaler vs.\ probe-without-scaler comparison whose accuracy gap is below $1\%$ across runs --- the input information is structurally destroyed rather than merely masked by the bias term.
  \item \emph{Monotonicity in $L$ at $r=\texttt{"all"}$:} ruled out by the $L$-sweep at the final epoch, where the most collapsed configuration is $L=2$ and configurations with $L=1$ or $L \ge 4$ all show milder bias inflation.
  \item \emph{Artefact of the $1$\,s tokeniser:} ruled out by two runs with $0.5$\,s patches at $r=\texttt{"all"}$ (\autoref{app:ablation-patch-length}), which display the full signature and score far below the healthy geometry downstream.
\end{itemize}

\subsection{Relation to prior work}\label{app:bic-related-work}

We are not aware of a prior description of bias-inflation collapse in its specific form --- a JEPA-only failure mode, on multichannel signals, triggered by a mask geometry that hides every channel of the same time-patch jointly, in which the encoder converges to a per-position lookup table $M(c, t)$ with an inflating bias $\mu$ and a vanishing input-driven residual $\delta$.
Several prior works, however, anticipate parts of the picture and we situate our contribution against them here.

\paragraph{Variance-fooling collapse.} \citet{liUnderstandingCollapseNoncontrastive2022} describe a partial dimensional collapse of non-contrastive Siamese SSL in which the per-dimension variance is preserved while the embedding distribution concentrates on a low-dimensional subspace; they show that PCA singular values detect the failure where per-dimension variance does not.
Bias-inflation collapse falls in the same broad family of ``high-variance, degenerate-representation'' regimes, but the failure mode is different: the encoder output is high-rank rather than low-rank (the effective rank actively grows from $\sim 5$ to $\sim 25$ during collapse, see \autoref{app:bic-vicreg}), and its degeneracy is position-indexed rather than subspace-restricted.

\paragraph{Mean drift in EMA-teacher SSL.} The DINO ``centering'' trick \citep{caronEmergingPropertiesSelfsupervised2021} subtracts a running mean of the teacher output to prevent one dimension from dominating the softmax, and is one of the earliest acknowledgements that EMA teachers can drift along their own mean.
\citet{moConnectingJointEmbeddingPredictive2024}, motivating C-JEPA, observe that I-JEPA ``inadequately learns the mean of patch representations'' and propose VICReg-style regularisation as a fix.
Bias-inflation collapse is consistent with this family of mean-drift concerns, but is sharper in two ways: the drift is position-indexed (different for each $(c, t)$ cell) rather than a single global mean vector, and the proposed VICReg fix is, by our analytical argument (\autoref{app:bic-vicreg}), structurally satisfied by the lookup-table solution.

\paragraph{Collapse via aggressive masking on highly correlated modalities.} \citet{baevskiData2vecGeneralFramework2022a} report that EMA-teacher latent prediction is prone to collapse for ``modalities where adjacent targets are very correlated and where longer spans need to be masked, e.g.\ speech'', and add target normalisation over the sequence (instance-norm for speech) to mitigate it.
Our $r=\texttt{"all"}$ regime is the multichannel-spatial analogue of their long-temporal-span observation: every channel of the same time-patch is masked jointly, so the only surviving context is on the temporal axis, which is itself highly autocorrelated for EEG.
data2vec's instance-norm fix targets a single sequence axis and does not generalise straightforwardly to a per-position $\mu(c, t)$ over a 2D channel-time grid.

\paragraph{Positional shortcuts in masked-prediction SSL.} \citet{zhangPCPMAELearningPredict2024} (PCP-MAE) document a ``center leakage'' phenomenon in point-cloud MAE: the decoder reconstructs plausibly even when fed only the positional embeddings of masked patches, without the encoder's output, indicating that positional metadata alone supplies the prediction signal.
\citet{barStochasticPositionalEmbeddings2024} show that deterministic positional embeddings let MIM models latch onto absolute location and propose stochastic positions to break the shortcut.
Bias-inflation collapse is a member of the same broader family ``the model exploits positional information to bypass the input'', but with a distinct mechanism: it is the encoder (not the decoder) that converges to a position-only function, and the JEPA student-teacher loss collapses to zero rather than the MAE reconstruction succeeding from positions alone.

\paragraph{JEPA loss as a poor proxy for representation quality.} A growing body of recent work argues that JEPA pre-training loss decreases independently of downstream representation quality, motivating non-loss probes such as RankMe \citep{garridoRankMeAssessingDownstream2023} or SIGReg \citep{balestrieroLeJEPAProvableScalable2025}.
Bias-inflation collapse is a concrete, mechanistically characterised instance of this broader concern, with one extra wrinkle: rank-based probes such as RankMe are themselves fooled by the lookup-table solution, since the per-position bias spreads isotropically across feature dimensions and yields a high effective rank (\autoref{app:bic-vicreg}).
Distributional regularisers that constrain the full embedding distribution, such as SIGReg, would in principle rule out the lookup-table fixed point, but we have not verified this empirically.

\section{Per-dataset breakdown of the downstream performance}\label{app:per-dataset-breakdown}

\begin{figure}[p]
  \centering
  \includegraphics[width=\linewidth]{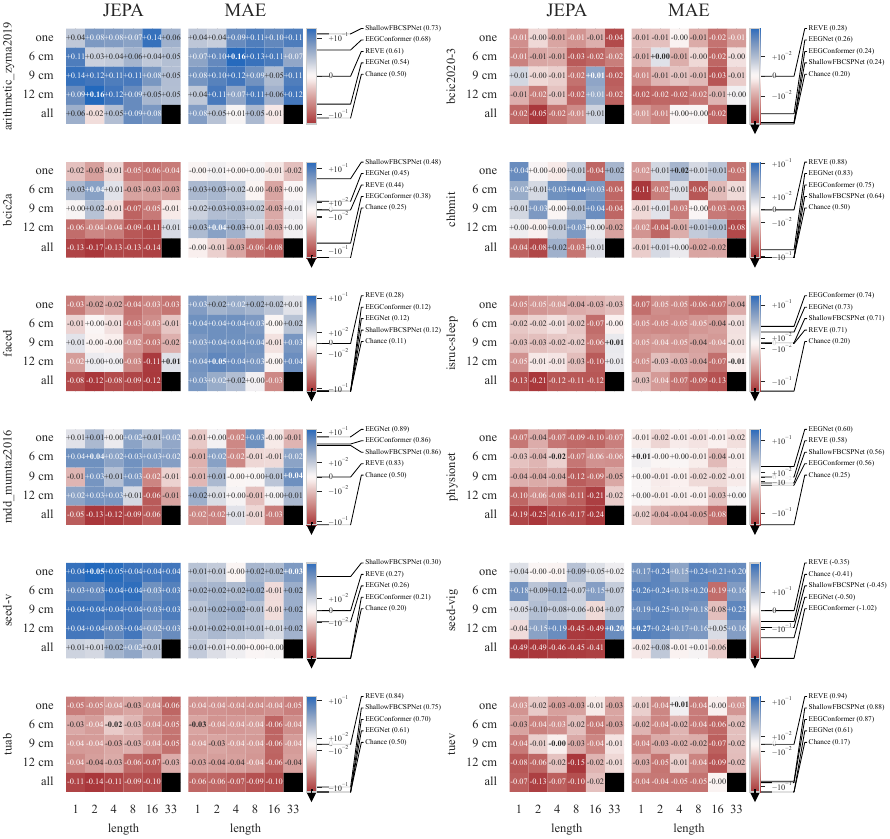}
  \caption{Per-dataset, per-(framework, $L$, $r$) score difference to
    the REVE ridge-probe baseline on OpenEEGBench. One row per
    dataset, one column per framework; each row uses its own diverging
    colour-bar so the diffs are interpretable in the dataset-native
    metric (balanced accuracy or $R^{2}$). Sidebars on the right of
    each row mark the absolute REVE level, the chance level, and the
    three supervised baselines (EEGNet, EEGConformer,
    ShallowFBCSPNet) trained end-to-end per task. The bottom row
    \emph{Average (norm.)} reproduces \autoref{fig:oeb-grid}\,(A) on
    the cross-dataset normalised scale.
    The trends summarised in Section~\ref{sec:results} are visible
    dataset-by-dataset: the moderate-$r$, short-$L$ cluster wins on
    most datasets, $r=\texttt{"all"}$ underperforms for JEPA on most
    datasets (with the exceptions of \texttt{arithmetic\_zyma2019}
    and \texttt{seed-v}), and $(L, r)=(2, 9\,\text{cm})$ is the
    aggregate JEPA optimum, although the local per-dataset peak
  sometimes lies elsewhere.}
  \label{fig:oeb-v9-per-dataset}
\end{figure}

\begin{figure}[p]
  \centering
  \includegraphics[width=.97\textwidth]{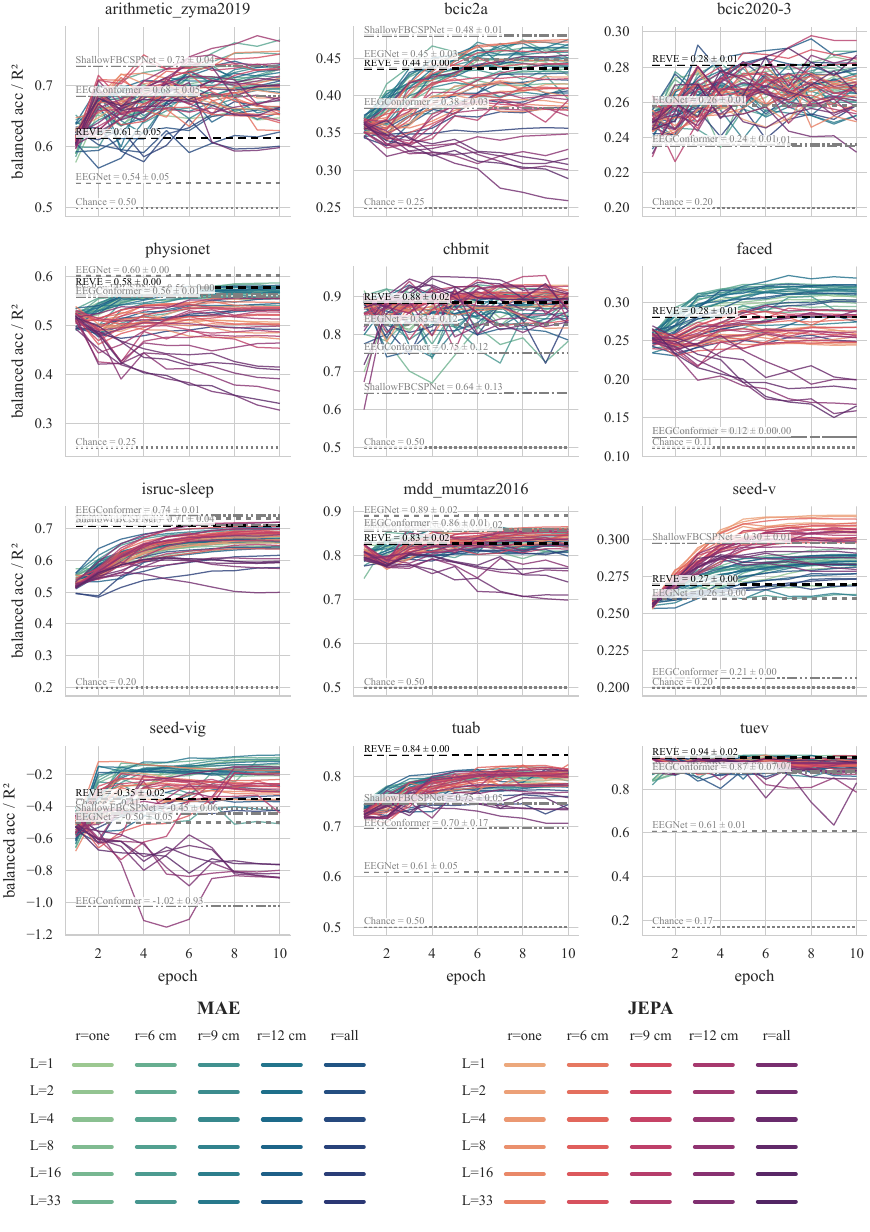}
  \caption{Per-dataset training trajectory of the
    downstream score across the $10$ pre-training epochs. Each
    line is one of the $58$ pre-training configurations
    (hue: framework), averaged across $3$ probe seeds.
    Per-dataset chance (dotted grey), the REVE baseline mean $\pm$ std
    (dashed black), and the three supervised baselines (EEGNet, EEG
    Conformer, ShallowFBCSPNet, dashed coloured) are overlaid for
    reference; the supervised baselines are trained end-to-end per
    task, while all FM curves use a frozen-feature ridge probe.
    Across datasets, both frameworks reach a near-final ranking by
    epoch $\sim\!5$; pathological JEPA runs at $r=\texttt{"all"}$ are
    visible as the bundle that drops below the REVE band early
  and never recovers.}
  \label{fig:oeb-timecourse-per-dataset}
\end{figure}

This appendix reports per-dataset versions of the aggregated panels of
\autoref{sec:results}. \autoref{fig:oeb-v9-per-dataset} expands the
average heatmap of \autoref{fig:oeb-grid}\,(A) into one row per
OpenEEGBench dataset, on the raw score scale (each row uses its own
  colour-bar so the per-dataset diff is interpretable in the
dataset-native metric, balanced accuracy or $R^{2}$). The bottom row
\emph{Average (norm.)} reproduces the across-dataset summary of
\autoref{fig:oeb-grid}\,(A) on the dataset-normalised scale and is
included for direct comparison with the main paper.
\autoref{fig:oeb-timecourse-per-dataset} likewise expands the
across-epoch summary of \autoref{fig:oeb-grid}\,(C) into one
sub-panel per dataset, with each $58$ pre-training run drawn as a
single transparent line so that within-framework variance over the
$10$ checkpoints is visible.

\clearpage
\section{Comparison with published EEG foundation models}\label{app:sota-comparison}

\autoref{tab:sota-comparison} compares our recommended geometry $(L, r) = (2\,\text{s}, 9\,\text{cm})$ with the public checkpoints of CBraMod~\citep{wangCBraModCrissCrossBrain2025}, EEGPT~\citep{wangEEGPTPretrainedTransformer2024}, LaBraM~\citep{jiangLargeBrainModel2024a} and BIOT~\citep{yangBIOTBiosignalTransformer2023} on the $12$ OpenEEGBench datasets.
All models are evaluated by linear probing on a frozen backbone, with one of two ways of fitting the linear layer.
The \emph{ridge} probe is the analytical fit used throughout this paper (\autoref{app:downstream-fm}); it was added to the benchmark after its original publication and requires the backbone to be entirely frozen.
The \emph{SGD} probe is the one of the original OpenEEGBench paper~\citep{guetschelOpenEEGBenchLiveCommunityDriven}, which fits the linear layer by iterative gradient descent; we report the scores published there, which cover ten of the twelve current datasets.
The ridge probe does not apply to EEGPT, BIOT and LaBraM, because they contain modules that must be retrained on each dataset and therefore cannot be fully frozen; these three models appear under the SGD probe only.

\begin{table}[!h]
  \centering
  \caption{Downstream scores of the recommended geometry against public EEG foundation models on the $12$ OpenEEGBench datasets (balanced accuracy; $R^2$ for \texttt{seed-vig}). Ridge: the analytical linear probe of this paper on fully frozen backbones. SGD: the linear probe of the original OpenEEGBench paper, whose published scores are reported (dashes: datasets it did not cover). Best score per row in bold.}
  \label{tab:sota-comparison}
  \footnotesize
  \setlength{\tabcolsep}{3.8pt}
  \begin{tabular}{lrrrrrrr}
    \toprule
    & \multicolumn{2}{c}{Ours $(2\,\text{s}, 9\,\text{cm})$, ridge} & \multicolumn{2}{c}{Public, ridge} & \multicolumn{3}{c}{Public, SGD (published)} \\
    \cmidrule(lr){2-3} \cmidrule(lr){4-5} \cmidrule(lr){6-8}
    Dataset & MAE & JEPA & REVE-Base & CBraMod & EEGPT & LaBraM & BIOT \\
    \midrule
    \texttt{arithmetic\_zyma2019} & $0.716$ & $\mathbf{0.737}$ & $0.614$ & $0.562$ & $0.551$ & $0.499$ & $0.553$ \\
    \texttt{bcic2020-3}       & $0.273$ & $0.277$ & $\mathbf{0.281}$ & $0.233$ & $0.207$ & $0.200$ & $0.216$ \\
    \texttt{bcic2a}           & $\mathbf{0.462}$ & $0.460$ & $0.436$ & $0.315$ & $0.296$ & $0.258$ & $0.264$ \\
    \texttt{chbmit}           & $0.888$ & $0.908$ & $0.882$ & $\mathbf{0.941}$ & --- & --- & --- \\
    \texttt{faced}            & $\mathbf{0.321}$ & $0.280$ & $0.281$ & $0.212$ & $0.174$ & $0.125$ & $0.155$ \\
    \texttt{isruc-sleep}      & $0.665$ & $0.682$ & $\mathbf{0.709}$ & $0.647$ & $0.487$ & $0.466$ & $0.702$ \\
    \texttt{mdd\_mumtaz2016}  & $0.842$ & $\mathbf{0.858}$ & $0.828$ & $0.776$ & $0.778$ & $0.783$ & $0.825$ \\
    \texttt{physionet}        & $0.568$ & $0.533$ & $\mathbf{0.577}$ & $0.474$ & $0.312$ & $0.294$ & $0.263$ \\
    \texttt{seed-v}           & $0.288$ & $\mathbf{0.307}$ & $0.269$ & $0.255$ & $0.212$ & $0.228$ & $0.239$ \\
    \texttt{seed-vig} ($R^2$) & $\mathbf{-0.108}$ & $-0.258$ & $-0.355$ & $-0.716$ & --- & --- & --- \\
    \texttt{tuab}             & $0.808$ & $0.805$ & $\mathbf{0.842}$ & $0.746$ & $0.715$ & $0.602$ & $0.573$ \\
    \texttt{tuev}             & $0.919$ & $0.930$ & $\mathbf{0.944}$ & $0.895$ & $0.338$ & $0.221$ & $0.357$ \\
    \bottomrule
  \end{tabular}
\end{table}

Under the ridge probe, our recommended geometry is ahead of CBraMod, the strongest published model after REVE-Base, on $11$ of the $12$ datasets, and none of the SGD-probed models match REVE-Base.
The models trained in this study are therefore not toy models: they are competitive with the published state of the art.

\section{Ablations}\label{app:ablations}

This section tests how far the conclusions of the core grid reach, along three axes: whether the recommended geometry transfers to other backbones (\autoref{app:ablation-backbones}), whether it survives changes of the mask ratio (\autoref{app:ablation-rho}), and whether bias-inflation collapse depends on the patch length (\autoref{app:ablation-patch-length}).
Throughout, downstream scores are mean normalised OpenEEGBench scores over the $12$ datasets and $5$ probe seeds, computed with the normalisation constants of the core grid (\autoref{app:score-normalisation}) so that they can be read against the grid.

\subsection{Transfer of the recommended geometry to other backbones}\label{app:ablation-backbones}

Does the recommended geometry help architectures other than REVE-Small?
To find out, we re-trained the CBraMod~\citep{wangCBraModCrissCrossBrain2025} and LaBraM~\citep{jiangLargeBrainModel2024a} architectures under MAE on our corpus, each with two masking strategies: the random masking they were originally trained with, $(r=\texttt{"one"}, L=1)$ at $\rho=0.5$, and our recommendation $(r=9\,\text{cm}, L=2)$ at the same $\rho=0.5$, for fairness.
One deviation from CBraMod deserves mention.
CBraMod encodes channel positions with 1D convolutions along the channel axis, which depend on the order in which the channels are stored; since that order is arbitrary, the encoding is flawed by design, and rather than re-implement it we used the sinusoidal encoding of the rest of the paper.
On both backbones, the recommended geometry improves the downstream score over the original strategy (\autoref{tab:ablation-backbones}), which supports the generalisability of our recommendation across architectures.

\begin{table}[!h]
  \centering
  \caption{Mean normalised OpenEEGBench score ($12$ datasets $\times$ $5$ probe seeds) of the CBraMod and LaBraM architectures re-trained under MAE on our corpus, with their published mask $(r=\texttt{"one"}, L=1)$ and with the recommended geometry $(r=9\,\text{cm}, L=2)$, both at $\rho=0.5$. The REVE-Small row is the same contrast in the core grid, at $\rho^\star=0.55$.}
  \label{tab:ablation-backbones}
  \footnotesize
  \begin{tabular}{lccc}
    \toprule
    Backbone & Published mask & Recommended geometry & Recommended better on \\
    \midrule
    REVE-Small (core grid) & $0.710$ & $0.800$ & $9$ of $12$ \\
    CBraMod                & $0.597$ & $0.640$ & $10$ of $12$ \\
    LaBraM                 & $0.605$ & $0.642$ & $7$ of $12$ \\
    \bottomrule
  \end{tabular}
\end{table}

\subsection{Sensitivity to the mask ratio}\label{app:ablation-rho}

The core grid fixes the mask ratio at $\rho^\star = 0.55$; this section asks whether its conclusions hold at other ratios.
Sweeping $\rho$ over the entire grid is out of reach ($58 \times 5 = 290$ pre-training runs), so we sweep $\rho \in \{0.25, 0.40, 0.55, 0.70, 0.85\}$ on the four geometries about which the paper makes its main claims, under both frameworks: the recommended cell $(r=9\,\text{cm}, L=2)$, random masking $(r=\texttt{"one"}, L=1)$, a large mask $(r=12\,\text{cm}, L=16)$ and the bias-inflation cell $(r=\texttt{"all"}, L=2)$.
This amounts to $30$ new pre-training runs; \autoref{tab:ablation-rho} reports their scores next to the corresponding cells of the core grid.

\begin{table}[!h]
  \centering
  \caption{Mean normalised OpenEEGBench score ($12$ datasets $\times$ $5$ probe seeds) as a function of the target mask ratio $\rho$, for four geometries under both frameworks. The $\rho=0.55$ column is the core grid.}
  \label{tab:ablation-rho}
  \footnotesize
  \begin{tabular}{llrrrrr}
    \toprule
    Framework & Geometry $(r, L)$ & $\rho=0.25$ & $0.40$ & $0.55$ (grid) & $0.70$ & $0.85$ \\
    \midrule
    MAE  & $(\texttt{"one"}, 1)$  & $0.59$  & $0.66$  & $0.71$ & $0.83$ & $0.78$ \\
    MAE  & $(9\,\text{cm}, 2)$    & $0.74$  & $0.82$  & $0.80$ & $0.81$ & $0.78$ \\
    MAE  & $(12\,\text{cm}, 16)$  & ---     & $0.57$  & $0.63$ & $0.59$ & $0.63$ \\
    MAE  & $(\texttt{"all"}, 2)$  & $0.61$  & $0.65$  & $0.65$ & $0.66$ & $0.56$ \\
    \midrule
    JEPA & $(\texttt{"one"}, 1)$  & $0.59$  & $0.64$  & $0.70$ & $0.72$ & $0.72$ \\
    JEPA & $(9\,\text{cm}, 2)$    & $0.79$  & $0.86$  & $0.83$ & $0.77$ & $0.73$ \\
    JEPA & $(12\,\text{cm}, 16)$  & ---     & $0.45$  & $0.46$ & $0.57$ & $0.59$ \\
    JEPA & $(\texttt{"all"}, 2)$  & $-0.34$ & $-0.16$ & $0.05$ & $0.32$ & $0.37$ \\
    \bottomrule
  \end{tabular}
\end{table}

The first lesson of \autoref{tab:ablation-rho} is that $\rho$ and geometry are not independent knobs: the same change of $\rho$ acts differently on different geometries.
The spatially large masks $(12\,\text{cm}, L=16)$ and $(\texttt{"all"}, L=2)$ perform poorly at every $\rho$ under both frameworks, which confirms that they are failure modes: their low scores come from the geometry, not from the ratio.
The bias-inflation cell (JEPA, $\texttt{"all"}$, $L=2$) is the cell that $\rho$ influences most: it gets worse at lower $\rho$ and better at higher $\rho$, while staying non-competitive throughout.
We have no explanation for this dependence at this point; it could be a lead for understanding the mechanism in a follow-up study.
The recommended geometry $(9\,\text{cm}, L=2)$ is the least sensitive to $\rho$.
A slightly lower ratio, $\rho=0.4$, gives a small gain over the default ($+0.023$ under MAE, $+0.030$ under JEPA), which suggests that this ratio might be optimal; we report it without leaning on it.
Random masking $(\texttt{"one"}, L=1)$, finally, degrades sharply at low $\rho$ but improves at $\rho=0.7$ under both frameworks, to the point of matching the recommended geometry under MAE.

This last observation has a geometric explanation.
Raising $\rho$ changes the effective geometry of the mask: the number of blocks $K$ grows, more blocks overlap, and the merged blocks become larger than the nominal geometry, all the more so when the nominal blocks are small.

The sweep also illustrates why the core grid holds $\rho$ fixed.
Changing $\rho$ changes the learning task itself: it sets how many tokens the network has to predict, hence over how many tokens the learning signal is spread, and how many unmasked tokens remain available to make the prediction; both factors affect the difficulty of the pretext task.
Were $\rho$ and the geometry $(L, r)$ to change together, these influences would combine, the task difficulty would move in an uncontrolled way, and the two effects could not be separated.
Published models use $\rho$ close to $0.55$, which we therefore took as a prior and held fixed.
The overall conclusion of the sweep is that the geometry recommendation $(9\,\text{cm}, L=2)$ holds.

\subsection{Patch length and bias-inflation collapse}\label{app:ablation-patch-length}

At $r=\texttt{"all"}$, the only context left to the predictor is the temporal axis, and the patch length sets how predictive that axis is: with shorter patches, within-channel temporal correlations are stronger and the lookup-table shortcut of \autoref{app:bic-mechanism} less attractive.
The collapse could therefore be an artefact of our $1$\,s patches, which this ablation tests.
The test cannot be made free of confounds.
The window duration is fixed at $30$\,s, so halving the patch length doubles the token count, and one of two things must give: (A)~keep the dimension of each token, so that the model size is unchanged but the signal is represented over twice as many features per second; or (B)~keep the total dimensionality, so that each token becomes smaller, which requires adapting the architecture to the new token size.
Either way the token count grows, and since attention is quadratic in it, a $0.5$\,s tokeniser is more expensive to pre-train than a $1$\,s one.
We nevertheless ran both options under JEPA at $(r=\texttt{"all"}, L=1)$ with $0.5$\,s patches; each run needed $4$~GPUs to complete in about the same time as the $1$\,s runs on $2$.
We then benchmarked the two runs downstream and re-ran the diagnostics of \autoref{app:bic-empirical} on them (\autoref{tab:ablation-patch-length}).
Both stay well below the healthy geometry on the downstream benchmark, and both display the complete collapse signature: $\|\mu\|$ inflates about twofold over pre-training, $\sigma_\text{token}$ stays flat or shrinks, the ratio $\sigma_\text{token}/\sigma_\text{chan}$ falls monotonically, and the intra-channel cosine reaches $0.94$--$0.98$.
Bias-inflation collapse is therefore not an artefact of the $1$\,s patch length: it persists at $0.5$\,s, and the shorter patch only makes pre-training more expensive.
The reference rows of \autoref{tab:ablation-patch-length} are cells of the core grid; for the reasons above, they are not directly comparable to the new runs.

\begin{table}[!h]
  \centering
  \caption{Patch-length ablation: diagnostics of \autoref{app:bic-empirical} at the final epoch, inflation of $\|\mu\|$ between epochs $1$ and $10$, parameters of the encoder / of the whole model, pre-training cost, and mean normalised OpenEEGBench score. Reference rows are cells of the core grid.}
  \label{tab:ablation-patch-length}
  \scriptsize
  \setlength{\tabcolsep}{3pt}
  \resizebox{\linewidth}{!}{%
  \begin{tabular}{lccccccccrc}
    \toprule
    Run & Patch & Framework & Token dim. & Mask $(r, L)$ & $\sigma_\text{token}/\sigma_\text{chan}$ & $\overline{\cos_\text{intra}}$ & $\|\mu\|$, ep.\ $1\!\to\!10$ & Params enc.\,/\,tot. & GPU-h & OEB score \\
    \midrule
    (A) same model size          & $0.5$\,s & JEPA & $512$ & $(\texttt{"all"}, 1)$   & $0.18$ / $0.17$ & $0.98$ / $0.98$ & $\times 2.0$ / $\times 2.2$ & $12.6$\,M / $21.3$\,M & $39.5$ & $0.598$ \\ %
    (B) same features per second & $0.5$\,s & JEPA & $256$ & $(\texttt{"all"}, 1)$   & $0.27$ / $0.24$ & $0.94$ / $0.95$ & $\times 2.1$ / $\times 2.2$ & $3.2$\,M / $5.3$\,M   & $30.6$ & $0.417$ \\ %
    \midrule
    Core grid, collapsed         & $1.0$\,s & JEPA & $512$ & $(\texttt{"all"}, 1)$   & $0.05$ / $0.08$ & $1.00$ / $1.00$ & $\times 2.0$ / $\times 2.0$ & $12.7$\,M / $21.3$\,M & $16.3$ & $0.320$ \\ %
    Core grid, collapsed         & $1.0$\,s & JEPA & $512$ & $(\texttt{"all"}, 2)$   & $0.02$ / $0.05$ & $1.00$ / $1.00$ & $\times 2.4$ / $\times 2.5$ & $12.7$\,M / $21.3$\,M & $14.0$ & $0.054$ \\ %
    Core grid, healthy           & $1.0$\,s & JEPA & $512$ & $(6\,\text{cm}, 2)$     & $0.47$ / $0.52$ & $0.87$ / $0.89$ & $\times 1.4$ / $\times 1.5$ & $12.7$\,M / $21.3$\,M & $14.0$ & $0.757$ \\ %
    Core grid, MAE control       & $1.0$\,s & MAE  & $512$ & $(\texttt{"all"}, 2)$   & $0.83$ / $0.95$ & $0.71$ / $0.72$ & $\times 1.0$ / $\times 1.0$ & $12.7$\,M / $21.2$\,M & $12.7$ & $0.652$ \\ %
    \bottomrule
  \end{tabular}}
\end{table}

\end{document}